\documentclass[pmlr]{jmlr}

\RequirePackage{graphicx}
 \usepackage{booktabs}
\usepackage{longtable}
\usepackage{pifont}

\usepackage{algorithmicx}
\usepackage{algorithm}
\usepackage[noend]{algpseudocode}
\usepackage{algorithm,algpseudocode}

\makeatletter
\def\set@curr@file#1{\def\@curr@file{#1}} 
\makeatother
\usepackage[load-configurations=version-1]{siunitx} 

\theorembodyfont{\upshape}
\theoremheaderfont{\scshape}
\theorempostheader{:}
\theoremsep{\newline}

\jmlrvolume{252}
\jmlryear{2024}
\jmlrworkshop{Machine Learning for Healthcare}

\title[Event-Based Contrastive Learning for Medical Time Series]{Event-Based Contrastive Learning for Medical Time Series}

\newcommand{\addrmit}{Massachusetts Institute of Technology, Cambridge, MA, USA}
\newcommand{\addrharvard}{Harvard Medical School, Boston, MA, USA}

\author{\Name{Nassim Oufattole}$^1$\thanks{Equal contribution.}
       \Email{nassim@mit.edu}
       \AND
       \Name{Hyewon Jeong}$^1\footnotemark[1]$
       \Email{hyewonj@mit.edu}
       \AND
       \Name{Matthew Mcdermott}$^2$
       \Email{matthew\_mcdermott@hms.harvard.edu}
       \AND
       \Name{Aparna Balagopalan}$^1$
       \Email{aparnab@mit.edu}
       \AND
       \Name{Bryan Jangeesingh}$^1$
       \Email{brytech@mit.edu}
       \AND
       \Name{Marzyeh Ghassemi}$^1$
       \Email{mghassem@mit.edu}
       \AND
       \Name{Collin Stultz}$^{1,2}$
       \Email{cmstultz@mit.edu}
        \AND
        $^1$ \addr \addrmit \\ 
        $^2$ \addr \addrharvard
} 

\begin{document}

\maketitle

\begin{abstract}
In clinical practice, one often needs to identify whether a patient is at high risk of adverse outcomes after some key medical event. For example, quantifying the risk of adverse outcomes after an acute cardiovascular event helps healthcare providers identify those patients at the highest risk of poor outcomes; i.e., patients who benefit from invasive therapies that can lower their risk. Assessing the risk of adverse outcomes, however, is challenging due to the complexity, variability, and heterogeneity of longitudinal medical data, especially for individuals suffering from chronic diseases like heart failure. In this paper, we introduce Event-Based Contrastive Learning (EBCL) - a method for learning embeddings of heterogeneous patient data that preserves temporal information before and after key index events. We demonstrate that EBCL can be used to construct models that yield improved performance on important downstream tasks relative to other pretraining methods. We develop and test the method using a cohort of heart failure patients obtained from a large hospital network and the publicly available MIMIC-IV dataset consisting of patients in an intensive care unit at a large tertiary care center. On both cohorts, EBCL pretraining yields models that are performant with respect to a number of downstream tasks, including mortality, hospital readmission, and length of stay.  In addition, unsupervised EBCL embeddings effectively cluster heart failure patients into subgroups with distinct outcomes, thereby providing information that helps identify new heart failure phenotypes.  The contrastive framework around the index event can be adapted to a wide array of time-series datasets and provides information that can be used to guide personalized care.
\end{abstract}
\section{Introduction}
\label{sec:intro}


Healthcare providers often aspire to identify a patient's risk of a future adverse event after some index event, such as an inpatient admission or intubation in the ICU. Prediction of such risks is important for the development of effective treatment strategies \citep{rahimi2014risk} and for ensuring that healthcare resources are allocated appropriately \citep{duong2021identification, jencks2009rehospitalizations}. This has motivated researchers to study a variety of learning algorithms, including directly supervised \citep{zhang2021graph, rajkomar2018scalable}, self-supervised \citep{tipirneni2022self, labach2023DuETT, jeong2023deep} and contrastive learning algorithms \citep{hyvarinen17a, agrawal2022leveraging} for identifying patients at the highest risk of adverse outcomes. 

Many contrastive learning methods (and more generally, representation learning methods) for medical time series data \citep{agrawal2022leveraging, tipirneni2022self, labach2023DuETT} strive to create latent representations that preserve temporal trends within the data. These approaches, however, generally ignore the fact that some portions of medical time series data are more informative than others. For example, data surrounding key medical events (e.g., an admission for a heart attack, or an admission associated with a new cancer diagnosis) are rich in information that plays a significant role in patient prognostication. 

Our approach, Event-Based Contrastive Learning (EBCL), diverges from existing work \citep{hyvarinen17a, agrawal2022leveraging, Dave_2022} by imposing a specialized pretraining contrastive loss solely on data around critical events, where the most clinically-relevant information regarding disease progression and prognosis is likely to be found. EBCL is a contrastive learning method, meaning it defines a latent space structure leveraging positive and negative pairs---where positive pairs of points should be close and negative pairs far apart in the latent space. In particular, we use patient data immediately before and after a key medical event as positive pairs, and data from different patients as negative pairs (Figure \ref{fig:ebcl}). The resulting embedding maps data surrounding key medical events to similar regions of the latent space, thereby encoding temporal trends.

We evaluate EBCL on two datasets: a private, multi-site cohort of heart failure patients and a cohort of intensive care unit (ICU) patients derived from MIMIC-IV \citep{johnson2023mimic}. Experiments are performed under a traditional pretraining/finetuning regime, in which model parameters are first initialized with EBCL pretraining and then specialized by task-specific finetuning. We perform empirical comparisons with published pretraining systems for the medical domain. \emph{We present finetuning results showing that EBCL consistently outperforms these baselines across all tasks and all events.} Furthermore, beyond traditional fine-tuning, we also show that EBCL embeddings produce informative representations of patient states. \emph{Linear probing with frozen EBCL embeddings on the heart failure cohort achieves AUCs at least $4.38$ points higher than other methods for predicting Mortality and $12.99$ points higher for predicting Length of Stay}. EBCL embeddings further yield highly expressive clusters that heart failure patients into distinct phenotypes, forming a basis for disease subtyping. With ablation studies, we demonstrate that the advantages of EBCL are uniquely due to its focus on temporal trends immediately surrounding key medical events. Finally, we reproduce these improvements on downstream tasks using mechanical ventilation and hypotension events in the ICU setting. 
Overall, our results strongly demonstrate the benefits of domain-specific latent space structure learning through its focus on key medical events and its unique temporal contrastive loss formulation. 
In summary, our contributions are as follows:
\begin{itemize}
    \item We propose a new contrastive pretraining method for time series data that encodes patient-specific temporal trends around key medical events in clinical data.
    \item We demonstrate that this formalism leads to improved predictive performance on downstream outcome prediction tasks, outperforming published pretraining baselines and supervised models.
    \item EBCL pretraining alone generates embeddings that are useful for identifying high-risk subgroups based on patient-specific temporal trends. This suggests these contrastive pretraining methods may be useful beyond downstream task prediction and for patient subtyping.
\end{itemize}

\subsection*{Generalizable Insights about Machine Learning in the Context of Healthcare}

\emph{Our results demonstrate that accounting for the clinical importance of key medical events explicitly during representation learning can significantly improve the quality of representations learned and eventual downstream performance, offering significant advantages over a variety of published baselines on several finetuning tasks over two clinical datasets.}
This simultaneously (1) underscores the utility of incorporating clinical domain knowledge into machine learning in healthcare, (2) demonstrates a beautifully simple approach to inject this domain knowledge during learning in a way that is both performant and efficient, and (3) produces models that can further better integrate with the realities of event-focused clinical workflows where key medical alerts or decision support tasks are often triggered by specific events in the clinical setting (e.g., prediction of readmission after discharge or of patient decompensation after negative cardiovascular events in intensive care).
We believe that not only can the success of our explicit pre-training strategy, EBCL, generalize to other medical datasets and tasks, but also that this mechanism of incorporating domain knowledge by better contrasting temporal dynamics before and after key areas of clinical change may be helpful for both future research into health AI and for the generation of models that can be more closely integrated into event-centric clinical decision support settings in practice.

\section{Related Works}

\paragraph{Representation Learning for Clinical Time-Series Data}
Medical time series datasets pose unique challenges due to their inherently high-dimensional and irregularly sampled nature, and the significant presence of missing data \citep{shukla2020survey}. 
These complexities demand focused strategies to encode \citep{tipirneni2022self} and represent such time-series data \citep{li2020learning, tipirneni2022self, lee2023learning, mcdermott2023event, labach2023DuETT} to enhance model performance, particularly in scenarios with limited data. Decision tree baselines, such as XGBoost \citep{chen2016xgboost} remain competitive for both tabular and tabular time series data \citep{labach2023DuETT, shwartzziv2021tabular}.  Many pretraining methods have also been explored, which include forecasting (STraTS) \citep{tipirneni2022self}  masked imputation (DuETT) \citep{mcdermott2021comprehensive, labach2023DuETT}, and weakly supervised methods \citep{mcdermott2021comprehensive}. More recently, contrastive learning methods \citep{agrawal2022leveraging, hyvarinen17a, pmlr-v225-king23a} have been applied to time series and multimodal temporal data, which is known to impose deeper structural constraints on latent space geometry \citep{mcdermott2023structure} motivating them as a better option for the pretraining objectives. We focus on advancing contrastive learning methods for time series data to impose and leverage meaningful latent space structures for downstream tasks.

\paragraph{Contrastive Learning for Medical Time-Series Data}
Contrastive learning is a form of self-supervised learning (SSL) that uses positive and negative pairs of instances \citep{franceschi2019unsupervised, tonekaboni2021unsupervised, eldele2021time, hyvarinen17a, agrawal2022leveraging, oord2018representation, kiyasseh2021clocs, raghu2023sequential, jeong2023deep}. Such techniques may use (1) automatically co-occurring sources of information \citep{liang2022mind, raghu2023sequential, zhang2022contrastive, heiliger2022beyond}, (2) augmented versions of the same information \citep{eldele2021time, oord2018representation,radford2021learning, kiyasseh2021clocs, li2022hi, raghu2023sequential, oh2022lead, gopal20213kg, cheng2020subject}, (3) similar data instances with smaller distances to one another \citep{jeong2023deep}, or (4) temporal proximity \citep{franceschi2019unsupervised, tonekaboni2021unsupervised} to generate positive and negative pairs. Compared to previous works defining views based on temporal proximity \citep{franceschi2019unsupervised, tonekaboni2021unsupervised}, EBCL defines views based both on domain-informed index events and temporal proximity around them.

\section{Event-Based Contrastive Learning (EBCL) for Medical Time-Series}
\label{sec:method}

\begin{figure*}[t]
\centering
  \includegraphics[width=\textwidth]{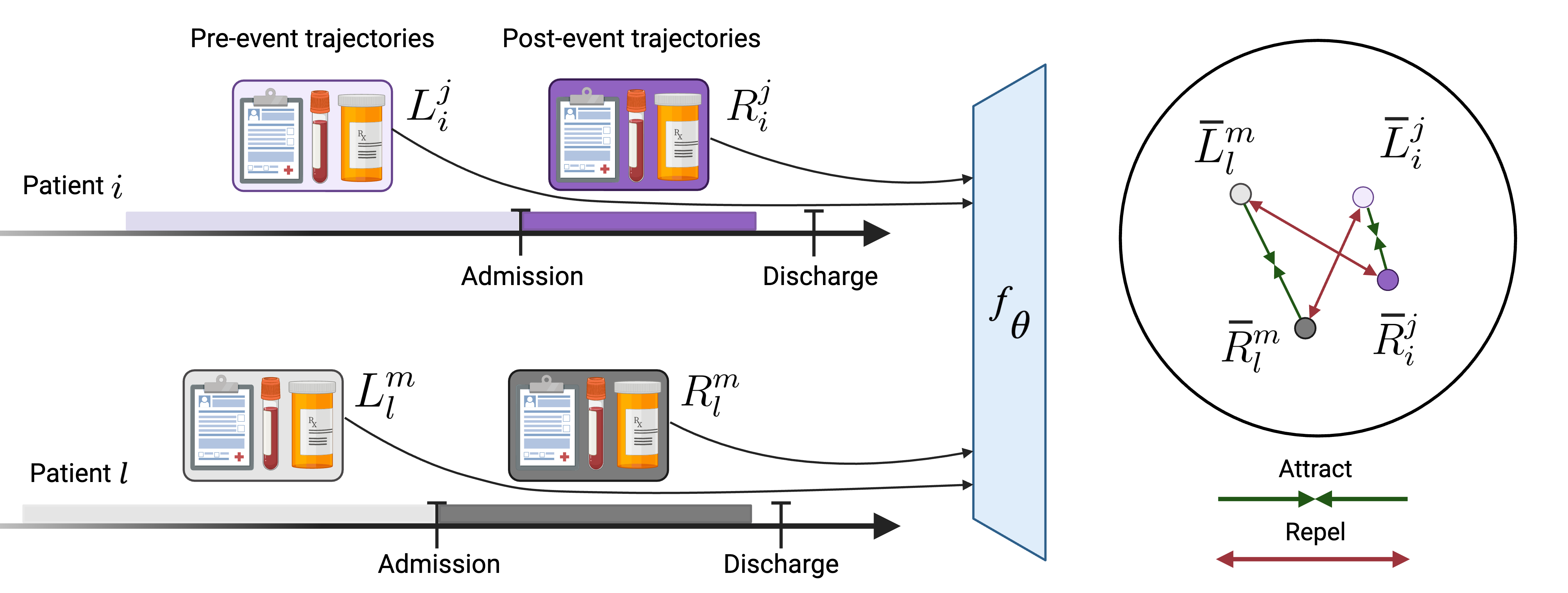}
  \caption{\textbf{Event-based contrastive learning (EBCL).} For patient $i$ and event $j$, we denote pre-event data, $L_i^j$, and post-event data, $R_i^j$. For EBCL pretraining, we sample a batch, $\mathcal{B}$, of pre and post-event trajectory pairs from a dataset, $D_{\mathcal{B}}=\{(L_i^j, R_i^j): i,j \in \mathcal{B}\}$. We choose the event of interest to be an inpatient admission. $L_i^j$ and $R_i^j$ are passed separately into a transformer encoder $f_\theta$ to get $\bar{L}_i^j = f_\theta(L_i^j)$ and $\bar{R}_i^j = f_\theta(R_i^j)$ which is pretrained with CLIP contrastive loss. The positive pairs are pre and post-event data of the same event, $(\bar{L}_i^j, \bar{R}_i^j)$. The negative pairs are mismatched pre-event and post-event trajectories from different patients, such as $(\bar{L}_i^j,\bar{R}_l^m)$ where $i \neq l$.}
 \label{fig:ebcl}
\end{figure*}

\subsection{Problem Formulation}
We use a formalism introduced in past work \citep{tipirneni2022self} to model heterogeneous time series data: Let $\mathcal{D}$ be a dataset containing $N$ patient trajectories, $\{X_1, \cdots, X_N\}$. A patient trajectory 
$X_i=[x_{i,1}, x_{i,2}, \ldots, x_{i,M_i}]$ is a chronologically ordered sequence of patient $i$'s $M_i \in \mathbb{N}$ observations. Each observation $x_{i, j}$ corresponds to an individual medical event (e.g., a laboratory test) and is encoded as a triple $x_{i,j} = (t_{i, j}, o_{i, j}, v_{i, j})$ where  $t_{i,j}$ is the time of the observation, $o_{i,j}$ denotes what medical result is being measured (e.g., what lab test is being observed), and $v_{i,j}$ is the actual value of the observation. For example, if a patient has a potassium test (encoded via $o = 7$) taken at time $t = \text{``12/1/2022, 11:30 a.m.''}$ which reports a value of 4.2 mEq/L, then their corresponding patient trajectory would contain a triple $(t_{i, j}, o_{i, j}, v_{i, j}) = (\text{``2/1/2024, 8:00 a.m.''}, 7, 4.2)$.   

To define the patient windows that EBCL will contrast during pre-training, we define the following. Given a window size $\tau \in \mathbb{N}$, let $t_{i, j}$ be the timestamp of a  clinical index event (e.g., a hospital admission) and define $L_i^j \subset X_i$ to be a subset of patient $i$'s trajectory consisting of the $\tau$ events prior to $t_{i, j}$ and $R_i^j \subset X_i$ be the subsequent $\tau$ events including and after $t_{i, j}$: 
\begin{align*}
    L_i^j &= [x_{i,j-\tau}, x_{i,j-\tau+1}, \ldots, x_{i, j-1}]\\
    R_i^j &= [x_{i,j}, x_{i, j+1}, \ldots, x_{i, j+\tau-1}].
\end{align*}

\subsection{Model Optimization Pipeline}
\label{sec:train}

We consider a two-stage learning problem where we pretrain a network $f_\theta$ and fine-tune on the outcome prediction tasks that are of interest. 

\paragraph{EBCL Pretraining} 


For pretraining a model, we sample an event and its corresponding pre-event and post-event dataset. We use a model, $f_\theta$ as an encoder, and get pre- and post-embeddings $f_\theta(L_i^j)$ and $f_\theta(R_i^j)$. We then compute the CLIP loss $\mathcal{L}\textsubscript{CLIP}$ \citep{radford2021learning} on a batch of these embeddings, where $(L_i^j, R_l^m)$ is a positive pair if $i=l$ and $j=m$. Intuitively, CLIP loss pushes together representations of positive pairs (i.e. paired left and right windows) and repels representations of negative pairs (i.e. mismatched left and right windows) as depicted in Figure \ref{fig:ebcl}.

\paragraph{Finetuning}  During fine-tuning, we use encoder $f_\theta$ to get representations finetuned for downstream outcome classification tasks. We use negative cross-entropy loss, $\mathcal{L}\textsubscript{CE}$. For tasks involving both pre- and post-event data (such as predicting 1-year mortality following hospital discharge, where 'pre' refers to data prior to admission and 'post' refers to data during admission but prior to discharge), we obtain embeddings of both ($f_\theta(L_i^j)$ and $f_\theta(R_i^j)$) and pass them through a shallow feedforward network (see Appendix Figure \ref{fig:architecture}) to arrive at a prediction for our label, $y_i^j$. For tasks that only use data prior to the key medical event (such as predicting if hospital length of stay will be greater than 1 week given only pre-admission event data), we pass only the pre-data embedding ($f_\theta(L_i^j)$) to the shallow feedforward network. EBCL weights after pretraining are used to initialize this model for the downstream tasks defined in Section \ref{sec:datasets}.  Furthermore, to prevent dataset leakage, we ensured the label $y_i^j$ of heart failure cohort (1-Year Mortality, 30-Day Readmission) to happen at least one day after the end of $R_i^j$. For predicting the length of stay (LOS) of the heart failure cohort and for predicting In-ICU Mortality and LOS of the MIMIC-IV cohort, we only use the pre-representation $f_\theta(L_i^j)$, and to avoid data leakage, outcomes must be at least one day after the end of $L_i^j$.

\section{Dataset and Methods}
\label{sec:experiments}

In this section, we introduce the dataset, outcome prediction tasks, and baseline models we used for the experiment.\footnote{Code used for the experiment is available at: \url{https://github.com/mit-ccrg/ebcl}} 

\subsection{Dataset and Tasks}
\label{sec:datasets}

We demonstrated our method using two datasets: a private dataset containing multi-site medical records of a heart failure cohort and the public MIMIC-IV ICU medical record dataset.

\subsubsection{Heart Failure Dataset} 
\paragraph{Dataset} We have assembled a multi-site cohort of $107,268$ patients with a prior diagnosis of heart failure. This cohort includes patient within a single hospital system with multiple locations (i.e. one hospital and its satellite locations). Collectively, this cohort had $383,254$ inpatient admissions, obtained from the electronic data warehouse of a large hospital network. The dataset includes patient trajectories over a maximum span of $40$ years and a maximum number of $3,275$ features, which includes labs, diagnoses, procedures, medications, tabular echocardiogram recordings, physical measurements (weight, height), and admissions/discharges. In our heart failure cohort described in Table \ref{tab:datasets}, we restrict our clinical events to inpatient admissions that have at least $16$ data points for both pre-admission and post-admission data. If a patient has no such event, they were not included in the cohort. We partitioned our compiled dataset into training (80\%), validation (10\%), and testing (10\%) with the split stratified so that no patients overlapped across splits. Additional information on dataset preprocessing is provided in the appendix Section \ref{apd:dataset}. 

\begin{table*}[hbtp]
  \caption{\textbf{Statistics for finetuning datasets from the electronic health record of Heart Failure cohort.}}
  \label{tab:datasets}
  \small
  \begin{center}
      \begin{tabular}{lccc}
  \toprule
    & \multicolumn{3}{c}{Heart Failure Cohort (Event: Inpatient Admission)}\\
    \cmidrule(r){2-4}
    Task & \# Patients & \# Events& \# Prevalence\\
    \midrule
    Readmission
    & 65,435
    & 262,734
    & 26.8\%
    \\
    Mortality 
    & 52,748
    & 195,747
    & 30.6\%
    \\
    LOS
    & 107,268
    & 383,254
    & 54.1\%
    \\
    \bottomrule
    \end{tabular}
    \end{center}
\end{table*}

\paragraph{Tasks} We finetune and evaluate on three binary downstream tasks: 30-day readmission, 1-year mortality, and 7-day LOS. The task of predicting 30-day readmission has been chosen for the heart failure cohort due to its critical importance for hospitals. This metric is financially significant, as hospitals face penalties under the Centers for Medicare \& Medicaid Services Readmission Reduction Program, which cost them over half a billion dollars in 2017 \citep{upadhyay2019readmission}. The datasets are summarized in Table \ref{tab:datasets}. Note that for the LOS task, we only use Pre-Admission data as input, as Post-Admission data would leak the LOS outcome. We also always restrict Post-Admission data, $R_n^i$, to the data prior to patient discharge, as this is the information that will be available at decision time for the 1-year mortality and 30-day readmission tasks. 

\subsubsection{MIMIC IV \citep{johnson2023mimic}}
\paragraph{Dataset and Events} MIMIC IV \citep{johnson2023mimic} is a public EHR dataset with ICU stay of patients admitted to Beth Israel Deaconess Medical Center between 2008 and 2019. We have assembled a cohort of patients who have hospital stay records and experienced the key events that frequently happen within the ICU (Table \ref{tab:mimic_dataset}). We utilize two medical events:
\begin{itemize}
    \item \textbf{Hypotension:} any time point where mean arterial pressure (invasive MAP or noninvasive NIMAP) transitions from over 60 to below 60 mmHg.
    \item \textbf{Mechanical Ventilation:} the start of a mechanical ventilation procedure.
\end{itemize}

\begin{table*}[hbtp]
\caption{\textbf{Statistics for finetuning datasets from the electronic health record of MIMIC-IV ICU cohort.}}
\label{tab:mimic_dataset}
\small
 \begin{center}
  \begin{tabular}{lccccc}
  \toprule
    & \multicolumn{5}{c}{MIMIC-IV ICU Cohort} \\
    \cmidrule(r){2-6}
    Task & \# Event-Type & \# Patients & \# Stays & \# Events & \# Prevalence\\
    \midrule
    Mortality
    & Hypotension
    & 35,234
    & 47,567
    & 342,884
    & 17.1\%
    \\
    LOS
    & Hypotension
    & 35,234
    & 47,567
    & 342,884
    & 48.1\%
    \\
    Mortality
    & Mechanical Ventilation
    & 23,269
    & 26,955
    & 31,420
    & 13.4\%
    \\
    LOS
    & Mechanical Ventilation
    & 23,269
    & 26,955
    & 31,420
    & 52.7\%
    \\
    \bottomrule
    \end{tabular}
    \end{center}
\end{table*}

\paragraph{Tasks} We finetune models on two binary outcome prediction tasks related to acute patient status in the ICU: In-ICU mortality and 3-day LOS, motivated by previous works \citep{Alghatani2021, nguyen2021clinical, Zhang2024, mcdermott2021comprehensive, wang2020mimic}. As post-event (hypotension, mechanical ventilation) could leak the outcome label as it could include the time horizon until the outcome, we limit the dataset to the dataset prior to the event for fine-tuning. 

\subsection{Model Architecture}
\label{sec:models}

For all experiments, unless specified otherwise, we use a transformer encoder as the backbone of our architecture (Figure \ref{fig:architecture}). The encoder has two encoder layers with the input of 512 observation sequences followed by a 128-dimension feed-forward layer between self-attention layers, and 32-dimension token embeddings. We then perform Fusion Self-Attention \citep{tipirneni2022self, ffattention}, by taking an attention-weighted average of the output embeddings of the transformer to get a single 32-dimension embedding. Finally, we have a linear projection to a 32-dimension embedding. Input sequences are required to have a length of at least 16 observations and are padded to a length of 512. Attention over padded tokens is masked in the transformer and the Fusion Self-Attention layer. This is the exact architecture from the paper \citep{tipirneni2022self} except we use a 128 dimension feed-forward layer instead of 2048 as this improved the supervised baselines in initial experiments.

\subsection{Baseline methods}
\label{sec:baselines}
To evaluate our method, we perform experiments with the following baselines and dataset preparation methods. For more information on methods, see Appendix Section \ref{apd:dataset}.
    
\paragraph{XGBoost} \citep{chen2016xgboost}: XGBoost is a tree boosting-based machine learning algorithm, widely used for classification and regression tasks with tabular datasets, and a competitive baseline for time-series prediction tasks \citep{mcdermott2023event}. 
    
\paragraph{Supervised Transformer (S-Trans)}: This corresponds to standard supervised training without EBCL  pretraining. The transformer model $f_\theta$ is initialized with random weights, and then the model is trained in a supervised fashion for a specific task.
    
\paragraph{Order Contrastive Pretraining (OCP)} \citep{agrawal2022leveraging}:
We take a continuous sequence of at most 512 tokens, split the sequence in half, and randomly swap the first and second halves (Appendix Figure \ref{fig:ocp}). We pretrain $f_\theta$ with the OCP objective for each patient $i$ where the pretraining task is to discriminate correct and switched sequencing. 
    
\paragraph{Self-supervised Transformer for Time-Series (STraTS)} \citep{tipirneni2022self}: STraTS represents transformer-based forecasting of time-series data. The transformer-based architecture they proposed is designed for handling sparse and irregularly sampled multivariate clinical time-series data. 

\paragraph{Dual Event Time Transformer (DuETT)} \citep{labach2023DuETT}: DuETT proposes a masked imputation pretraining task for detecting the presence of a feature and its value.

\subsection{Finetuning and Downstream Outcome Prediction}
We finetune the pretrained models (OCP, STraTS, DuETT, and EBCL) with a single fully connected layer to predict outcomes. We perform an extensive learning rate and dropout hyperparameter search for pretraining and finetuning and use a maximum of 300 epochs for pretraining and 100 epochs for finetuning. More details are provided in the Appendix Section \ref{apd:tune}. We take the epoch with the highest validation set performance for pretraining and finetuning. We pretrain our models and run 5 random seeds for finetuning and report the mean and standard deviation of results across these seeds.
\section{Experiments and Results}

We present the following key results: 1) EBCL outperforms all baselines on fine-tuning tasks. 2) EBCL yields richer embeddings than competitor pretraining models, as assessed on the heart failure cohort. 3) The definition of the domain-informed event and its closeness to the sampled event are key to our performance gains.

\subsection{EBCL Outperforms All Baselines in the Heart Failure Dataset}
In addition to EBCL, we consider several baseline methods for comparison. XGBoost builds a representation that summarizes information along the long period of clinical time series. OCP \citep{agrawal2022leveraging} learns features that are sensitive to temporal reversal (which are called \textit{least time reversible features}). STraTS \citep{tipirneni2022self} is designed to learn features that are useful for forecast observations during inpatient stays, which helps build a representation that captures relevant features and temporal dependencies. DuETT \citep{labach2023DuETT} representation learns missingness-invariant representations of data from masked imputation for accurate downstream prediction robust to missingness of input pretraining data. Each time-series pretraining baseline model learns distinct features that, in principle, help the model perform better for subsequent finetuning clinical outcome classification tasks. By contrast, EBCL learns patient-specific temporal trends associated with specific index medical events.  

\begin{table*}[h!]
\caption{\textbf{EBCL Pretraining improves results over a supervised baseline and time-series pretraining baselines in the Heart Failure Dataset.} For the heart failure cohort, we summarize the downstream finetuning performance using both the area under the receiver operating characteristic (AUC) of three prediction tasks (30-Day Readmission, 1-Year Mortality, and 1-Week Length of Stay (LOS)) averaged over 5 runs with different seeds. We present finetuning performance on the MIMIC-IV Dataset for the outcomes of in-ICU Mortality and 3-Day LOS. Across all tasks, EBCL results (boldfaced) were statistically significantly better than all other tested models.
}
\label{tab:pred}
\small
\begin{center}
  \begin{tabular}{lccccccc}
  \toprule
    & \multicolumn{3}{c}{Heart Failure Cohort} \\
    \cmidrule(r){2-4}
    & 30-Day Readmission
    & 1-Year Mortality
    & 1-Week LOS 
    \\
    \midrule
    \midrule
    XGBoost
    & 70.85 $\pm$ 0.08
    & 80.33 $\pm$ 0.19
    & 79.74 $\pm$ 0.10
    \\
    S-Trans
    & 70.28 $\pm$ 0.16
    & 81.54 $\pm$ 0.24
    & 88.74 $\pm$ 0.48
    \\
    \midrule
    OCP 
    & 70.27 $\pm$ 0.32
    & 80.06 $\pm$ 0.29
    & 90.06 $\pm$ 0.25
    \\
    STraTS 
    & 70.06 $\pm$ 0.11
    & 79.95 $\pm$ 0.62
    & 88.36 $\pm$ 1.09
    \\
    DueTT 
    & 69.51 $\pm$ 0.50
    & 79.39 $\pm$ 0.16
    & 75.35 $\pm$ 0.61 
    \\
    \midrule
    EBCL
    & \textbf{71.66 $\pm$ 0.03}
    & \textbf{82.43 $\pm$ 0.07}
    & \textbf{90.98 $\pm$ 0.05}
    \\
    \bottomrule
\end{tabular}
\end{center}
\end{table*}

EBCL achieves a significant improvement over all baseline models for three predictive tasks (1-Year Mortality, 30-Day Readmission, 1-Week LOS) in the heart failure dataset (Table \ref{tab:pred}, \ref{tab:pred_apr}). We note that pretraining methods can often achieve improved performance relative to a supervised baseline when the pretraining dataset is significantly larger than the finetuning datasets \citep{bert, vit}. However, in this case, EBCL achieves better performance over the supervised baseline even though the pretraining and finetuning datasets are of similar size. Consequently, the improvement in performance arises from solving the contrastive learning task and not the fact that pretraining leverages a larger dataset. Contrasting around our index event better captures temporal trends than other contrastive and generative pretraining methods.

\subsection{EBCL Embeddings are More Informative than Other Baselines}

\paragraph{Linear Probing} To gauge the extent to which the pretrained embeddings contain important information that can be leveraged for downstream predictive tasks, we employed a linear probing evaluation on the heart failure cohort. This involved fitting a logistic regression classifier on the frozen embeddings generated by the model. The coefficient for L2 regularization was tuned. We compute the mean and standard deviation of logistic regression results over 5 seeds of pretraining each method, and results are reported in Table \ref{tab:linear_probe} and \ref{tab:linear_probe_apr}.

\begin{table*}[h!]
\caption{\textbf{Linear Probing for Evaluating the Pretrained Embeddings.} EBCL (boldfaced) yielded the embedding that performs the best across all tasks.}
\small
\label{tab:linear_probe}
\begin{center}
  \begin{tabular}{lccc}
  \toprule
    & 30-Day Readmission
    & 1-Year Mortality
    & 1-Week LOS
    \\
    \midrule
    OCP 
& 65.04 ± 0.39 & 70.34 ± 0.83 & 58.75 ± 0.23
    \\
    STraTS 
& 60.88 ± 0.75 & 65.33 ± 0.99 & 57.37 ± 0.96
    \\ 
    DuETT 
& 58.39 ± 3.61 & 63.52 ± 6.71 & 55.29 ± 1.61
    \\ 
    \midrule
    EBCL
& \textbf{65.40 ± 0.16} & \textbf{74.72 ± 0.36} & \textbf{71.74 ± 2.08}
    \\
\bottomrule
\end{tabular}
\end{center}


\end{table*}

Frozen EBCL representations consistently outperformed other baseline methods for all three tasks evaluated (Table \ref{tab:linear_probe}, \ref{tab:linear_probe_apr}). We additionally see dramatic improvements when using a KNN classifier (results and methodology are in Appendix Section \ref{app:knn}). This implies that neighbors within the EBCL latent space are more similar in outcomes than those derived from any other baseline model, which indicates that EBCL inherently realizes a latent space that naturally stratifies outcomes.


\paragraph{Heart Failure Outcome Subtyping with EBCL Representations}
\begin{figure*}[h!]
\centering
  \includegraphics[width=\textwidth]{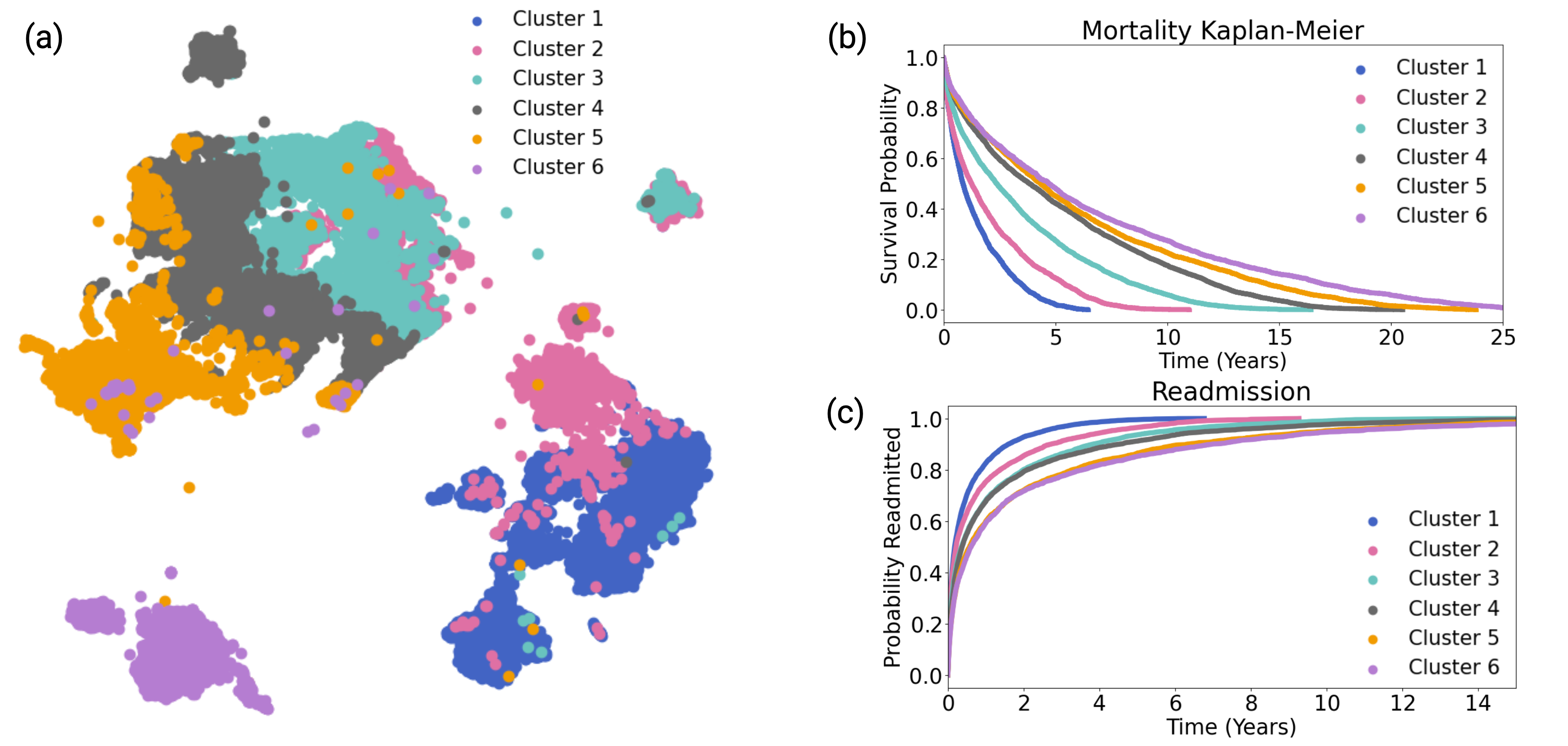}
  \caption{\textbf{Clustering and outcome risk subtyping from pretrained EBCL Embedding.} \textbf{(a)} Clustering and dimensionality reduction of pretrained embedding showed distinct heart failure clusters with unique outcome prognoses. Clusters are sorted by the prevalence of 1-Year Mortality (cluster 1 is the fatal group with the highest mortality rate, while cluster 6 is the healthiest group). The survival curve of the identified clusters plotting \textbf{(b)} time to mortality and \textbf{(c)} time to readmission}
  \label{fig:cluster}
\end{figure*}
We further explore the extent to which the learned embedding space naturally stratifies patient outcomes of interest, and find that EBCL embeddings are highly indicative of patient outcomes. In particular, using $K$-means clustering with $K=6$ (determined using elbow method as described in Appendix \ref{apd:cluster}), we see that EBCL clusters correspond to distinct heart failure patient subgroups. We compare the distributions of time-to-next-readmission and time-to-mortality (Figure \ref{fig:cluster} (b) and (c)), and using a $t$-test with p-value threshold $0.01$, we find a significant difference between the time-to-outcome distributions of patients in any two clusters. These relationships also hold on the outcome prevalence of each cluster. The cluster $1$ had the highest risk of 1-year mortality ($53.9$\%) and $30$-day readmission ($34.1$\%) rate; in contrast, patients in cluster $6$ had a more favorable prognosis with a lower risk of 1-year mortality ($21.0$\%) and 30-day readmission ($18.8$\%).

\begin{table*}[h!]
\caption{\textbf{EBCL Pretraining improves downstream outcome prediction over a supervised baseline and time-series pretraining baselines in the MIMIC-IV cohort.} We summarize the finetuning performance using two metrics: area under the receiver operating characteristic (AUC) of two prediction tasks (In-ICU Mortality and 3-Days Length of Stay (LOS)). The result has averaged over 5 runs with different seeds.  
}
\label{tab:mimic_pred}
\small
\begin{center}
  \begin{tabular}{lccccccc}
  \toprule
    &\multicolumn{2}{c}{MIMIC-IV Hypotension} &\multicolumn{2}{c}{MIMIC-IV Mechanical Ventilation}\\
    \cmidrule(r){2-3}\cmidrule(r){4-5}
    & In-ICU Mortality & 3-Day LOS & In-ICU Mortality & 3-Day LOS \\
    \midrule
    \midrule
    XGBoost
    & 80.07 $\pm$ 0.12
    & 80.19 $\pm$ 0.20
    & 81.45 $\pm$ 0.05
    & 77.96 $\pm$ 0.05
    \\
    S-Trans
    & 81.92 $\pm$ 0.32
    & 80.46 $\pm$ 0.14
    & 83.65 $\pm$ 10.31
    & 80.55 $\pm$ 0.10
    \\
    \midrule
    OCP 
    & 81.89 $\pm$ 0.21
    & 80.34 $\pm$ 0.09
    & 88.99 $\pm$ 0.18
    & 80.51 $\pm$ 0.13
    \\
    STraTS 
    & 82.73 $\pm$ 0.20
    & 80.46 $\pm$ 0.06
    & 88.99 $\pm$ 0.18
    & 80.51 $\pm$ 0.13
    \\
    DueTT 
    & 64.94 $\pm$ 1.75
    & 73.33 $\pm$ 2.47
    & 52.00 $\pm$ 0.71
    & 48.28 $\pm$ 10.04
    \\
    \midrule
    EBCL
    & \textbf{83.02 $\pm$ 0.08}
    & \textbf{80.70 $\pm$ 0.03}
    & 89.20 $\pm$ 0.35
    & 81.36 $\pm$ 0.05
    \\
    \bottomrule
\end{tabular}
\end{center}

\caption{\textbf{Linear Probing for Evaluating the Pretrained Embeddings.} EBCL (boldfaced) yielded the embedding that performs the best across all tasks.}

\label{tab:mimic_linear_probe}
\small
\begin{center}
  \begin{tabular}{lccccccc}
  \toprule
    &\multicolumn{2}{c}{MIMIC-IV Hypotension} &\multicolumn{2}{c}{MIMIC-IV Mechanical Ventilation}\\
    \cmidrule(r){2-3}\cmidrule(r){4-5}
    & In-ICU Mortality & 3-Day LOS & In-ICU Mortality & 3-Day LOS \\
    \midrule
    \midrule
    OCP 
    & 69.86 ± 0.67 & 72.13 ± 0.36 & 68.99 ± 1.92 & 68.25 ± 0.41
    \\
    STraTS 
    & 72.13 ± 1.36 & 73.06 ± 1.11 & 73.76 ± 0.64 & 69.58 ± 0.39
    \\
    DueTT 
    & 72.37 ± 1.29 & 74.94 ± 0.71 & 51.29 ± 0.00 & 58.16 ± 0.01
    \\
    \midrule
    EBCL
    & \textbf{77.80 ± 0.44} & \textbf{77.95 ± 0.43} & \textbf{79.26 ± 1.27} & \textbf{75.85 ± 0.63}
    \\
    \bottomrule
\end{tabular}
\end{center}
\end{table*}

\subsection{EBCL Generalizes to Acute Patient Cohort of MIMIC-IV ICU}
The applications above (Table \ref{tab:pred}, \ref{tab:linear_probe}) focus on applications of EBCL in a large heart failure cohort derived from several tertiary care centers, where the index event was hospital admission (Figure \ref{fig:ebcl}). To determine whether the approach is fruitful in other cohorts, and with other types of index events, we applied EBCL to the MIMIC-IV cohort \citep{johnson2023mimic}, which consists of patients admitted to an ICU at a major tertiary care center. We explored two key medical events: 1) hypotensive episodes (defined as a mean arterial pressure $<$ 60mmHg) and 2) mechanical ventilation. The outcomes of interest included In-ICU mortality and 3-day length of ICU stay. Across all tasks and events, EBCL yields the highest AUC for both finetuning (Table \ref{tab:mimic_pred}, \ref{tab:mimic_pred_apr}) and linear probing (Table \ref{tab:mimic_linear_probe}, \ref{tab:mimic_linear_probe_apr}) for predicting outcomes, compared with the baselines.

\subsection{Ablation Studies}

\paragraph{Performance Gains Observed with EBCL were Uniquely due to Key Event and Closeness of Data}
The novelty of our method arises from 1) leveraging the clinically important event and 2) using the data around the index event. We perform a series of ablation studies on the heart failure cohort to analyze the effect of defining the event and sampling observations around the event. 

\paragraph{Effect of the definition of event} We evaluate the importance of selecting a clinically significant event for pretraining by selecting non-inpatient admission events. We first use any non-inpatient visits as the EBCL index event (such as outpatient visits or emergency unit visits) instead of inpatient admission (\textbf{Non-Adm EBCL}). Our standard EBCL model, which uses inpatient hospital admission as the index event, outperformed an approach that specifically excludes inpatient admissions from the event set (Table \ref{tab:ablation}). 

As the Non-Adm EBCL experiments include encounters with the emergency room, we also performed a second set of experiments where events were restricted to only outpatient encounters (thereby excluding all emergency department encounters). These \textbf{Outpatient EBCL} experiments also underperformed standard EBCL. These results highlight the critical role of inpatient admissions in learning consistent patterns for predicting clinically significant outcomes, suggesting that the context around inpatient events is essential for optimal model performance in this cohort.

\begin{table*}[ht!]
\caption{\textbf{Ablation Studies}. We compare the EBCL variants by applying various definitions of events and different sampling strategies. We summarize the downstream finetuning performance (AUROC) of three prediction tasks averaged over 5 runs with different seeds, where we boldface statistically significant results. *Note that FT stands for finetuning data and the entry both means both pre and post data are used, but for predicting 1-Week LOS we always only used the Pre-event dataset to avoid data leakage.}
\label{tab:ablation}
\begin{center}
\small
\centering
  \begin{tabular}{lllcccccc}
  \toprule
    & Index Event & FT* & 30-Day Readmission & 1-Year Mortality & 1-Week LOS \\
    \midrule
    EBCL
    & Inpatient
    & Both
    & \textbf{71.66 $\pm$ 0.03}
    & \textbf{82.43 $\pm$ 0.07}
    & \textbf{90.98 $\pm$ 0.05}
    \\
    Censoring
    & Inpatient 
    & Both
    & 70.63	$\pm$ 0.10
    & 81.35	$\pm$ 0.04
    & 90.43 $\pm$ 0.11
    \\
    Non-Adm 
    & Non-Inpatient
    & Both
    & 70.73 $\pm$ 0.11
    & 81.89	$\pm$ 0.10
    & 90.12	$\pm$ 0.17
    \\
    Outpatient
    & Outpatient
    & Both
    & 70.51	$\pm$ 0.14
    & 81.80 $\pm$ 0.03
    & 89.68 $\pm$ 0.14
    \\
    \midrule
    Pre Event
    & Inpatient
    & Pre
    & 70.48 $\pm$ 0.09
    & 79.63 $\pm$ 0.06
    & \ding{55}
    \\
    Post Event
    & Inpatient
    & Post
    & 69.20 $\pm$ 0.09
    & 80.36 $\pm$ 0.03
    & \ding{55}
    \\
    \midrule
\end{tabular}
\end{center}
\end{table*}

\paragraph{Effect of Censoring Observations} We evaluate the importance of sampling observations locally around the index event by introducing a censoring window around the EBCL inpatient admission index event (\textbf{EBCL with Censoring}). In this experiment, we sample consecutive input observation points that are away from the index event by setting the censoring window (Appendix Figure \ref{fig:censoring}). Pre-event observations were chosen from the window preceding the censoring window before the index event. Post-event observations were chosen from the window following the censoring window after the index event. EBCL with Censoring demonstrated lower performance compared to the standard EBCL (Table \ref{tab:ablation}). This outcome reinforces the fundamental advantage of EBCL, which capitalizes on the proximity of data around the index event. Hence, leveraging data close to the index event is crucial for the model's effectiveness in clinical prediction tasks.


\section{Discussion}
\label{sec:conclusion}


\subsection{EBCL pretraining improves Outcome Stratification in Temporally Rich Problem Setting}

Our experimental results on both heart failure and MIMIC-IV ICU patient records show the generalizability of our method on different clinical time-series datasets. This experiment was meaningful in demonstrating the effectiveness of defining medically critical tasks for specific cohorts and leveraging time-series datasets surrounding them that are useful for downstream risk stratification of critically ill patient cohorts. Notably, EBCL showed superiority over other methods in capturing patient status, which is emphasized by improved clustering. These results underscore the robustness of EBCL in handling complex clinical time-series data.

\subsection{Event Centricity Aligned with Clinical Workflow Contributes to Improved Risk Stratification}

Our method is most appropriate in settings where we have temporal sequences of datasets and some knowledge about what types of events are likely to be clinically most meaningful. We demonstrated our method using event-centric downstream outcome prediction tasks for acute and chronic patient care scenarios. The primary use case, exemplified in this paper, is to predict cardiovascular outcomes specifically for the heart failure cohort using the in-hospital time-series dataset. Heart failure provides a particularly good application domain because the trajectory of patients with heart failure is punctuated by frequent hospital admissions, where each admission is associated with a further decrease in myocardial function \citep{Gheorghiade2005}; i.e., hospital admissions are particularly important in the health trajectory of these patients. For this application, we relied on prior domain knowledge about heart failure in general and its pathophysiology to identify what types of clinical events are likely to be most impactful. 

For the domain-informed events of the ICU dataset, we used the two most prevalent events corresponding to ICU stay: hypotension and mechanical ventilation. However, depending on the cohort of interest and the downstream task of interest the key event could include other medically important events like acidosis, alkalosis, sepsis, and hypoxia. 

The identification of relevant `events' to guide EBCL pretraining need not be restricted to hospital admissions. An event, for example, can encompass different clinical occurrences, including episodes of hypoglycemia in diabetic patients, a new diagnosis of hypertension made at an outpatient visit, etc. The key insight is that a key medical event is one where important clinical features change and/or new trends in the patient's clinical trajectory are expected. EBCL provides event-centric learning of health data which reflects the real-world clinical workflows. By doing so, EBCL not only captures the essential dynamics of patient care but also significantly improves risk stratification, providing a robust foundation for critical healthcare decisions.


\subsection{Limitations and Future Work} 
There are several limitations and areas for future work that EBCL inspires. Firstly, the effective application of the EBCL requires preliminary domain knowledge to identify suitable events for pretraining. This limits the applicability of EBCL in settings in which domain knowledge is not available, or in which finetuning tasks are not related to the events chosen for use during pre-training. Adapting EBCL to work in settings both where domain knowledge is either unavailable or partially available and in areas where finetuning tasks require more flexibility in how they relate to the key medical events used in pre-training are two areas of future work that could thus extend EBCL's generalizability to different clinical settings. Another area of future work could be to extend EBCL's contrastive loss framework to incorporate additional notions of similarity between patients or portions of patient records in addition to colocation around key medical events, such as future patient diagnoses or adverse events. This could enrich the geometry in the learned representation space by EBCL and capture deeper, more clinically meaningful relationships. Finally, the EBCL pretraining objective can be extended to additional modalities of health data with or multimodal health datasets.

\section{Conclusion}
EBCL, a novel pretraining scheme for medical time series data, learns patient-specific temporal representations around clinically significant events. We demonstrate that the method outperforms previous contrastive, generative, and supervised baselines on 3 different finetuning tasks over two clinical datasets. We further show that EBCL pretraining generates a rich latent space characterized by: 1) significant improvements in classification performance with linear probing on EBCL latent space and 2) improvement in identifying high risk patient subgroups using embeddings arising from EBCL pretraining alone. From a set of ablation studies, we show that the key to performance gains in EBCL is from the introduction of the event and the nearness of data around that index event. These results demonstrate key insights about representation learning over medical record data that underscore the importance of integrating domain knowledge and provide a translatable vehicle to do so, which will improve and extend the state of research in health AI overall.

\bibliography{references}

\begin{thebibliography}{48}
\providecommand{\natexlab}[1]{#1}
\providecommand{\url}[1]{\texttt{#1}}
\expandafter\ifx\csname urlstyle\endcsname\relax
  \providecommand{\doi}[1]{doi: #1}\else
  \providecommand{\doi}{doi: \begingroup \urlstyle{rm}\Url}\fi

\bibitem[Agrawal et~al.(2022)Agrawal, Lang, Offin, Gazit, and
  Sontag]{agrawal2022leveraging}
Monica~N Agrawal, Hunter Lang, Michael Offin, Lior Gazit, and David Sontag.
\newblock Leveraging time irreversibility with order-contrastive pre-training.
\newblock In \emph{International Conference on Artificial Intelligence and
  Statistics}, pages 2330--2353. PMLR, 2022.

\bibitem[Alghatani et~al.(2021)Alghatani, Ammar, Rezgui, and
  Shaban-Nejad]{Alghatani2021}
Khalid Alghatani, Nariman Ammar, Abdelmounaam Rezgui, and Arash Shaban-Nejad.
\newblock Predicting intensive care unit length of stay and mortality using
  patient vital signs: Machine learning model development and validation.
\newblock \emph{JMIR Medical Informatics}, 9\penalty0 (5):\penalty0 e21347, May
  2021.
\newblock ISSN 2291-9694.
\newblock \doi{10.2196/21347}.
\newblock URL \url{http://dx.doi.org/10.2196/21347}.

\bibitem[Chen and Guestrin(2016)]{chen2016xgboost}
Tianqi Chen and Carlos Guestrin.
\newblock Xgboost: A scalable tree boosting system.
\newblock In \emph{Proceedings of the 22nd acm sigkdd international conference
  on knowledge discovery and data mining}, pages 785--794, 2016.

\bibitem[Cheng et~al.(2020)Cheng, Goh, Dogrusoz, Tuzel, and
  Azemi]{cheng2020subject}
Joseph~Y Cheng, Hanlin Goh, Kaan Dogrusoz, Oncel Tuzel, and Erdrin Azemi.
\newblock Subject-aware contrastive learning for biosignals.
\newblock \emph{arXiv preprint arXiv:2007.04871}, 2020.

\bibitem[Dave et~al.(2022)Dave, Gupta, Rizve, and Shah]{Dave_2022}
Ishan Dave, Rohit Gupta, Mamshad~Nayeem Rizve, and Mubarak Shah.
\newblock {TCLR}: Temporal contrastive learning for video representation.
\newblock \emph{Computer Vision and Image Understanding}, 219:\penalty0 103406,
  jun 2022.
\newblock \doi{10.1016/j.cviu.2022.103406}.
\newblock URL \url{https://doi.org/10.1016%2Fj.cviu.2022.103406}.

\bibitem[Devlin et~al.(2019)Devlin, Chang, Lee, and Toutanova]{bert}
Jacob Devlin, Ming-Wei Chang, Kenton Lee, and Kristina Toutanova.
\newblock Bert: Pre-training of deep bidirectional transformers for language
  understanding, 2019.

\bibitem[Dosovitskiy et~al.(2020)Dosovitskiy, Beyer, Kolesnikov, Weissenborn,
  Zhai, Unterthiner, Dehghani, Minderer, Heigold, Gelly, et~al.]{vit}
Alexey Dosovitskiy, Lucas Beyer, Alexander Kolesnikov, Dirk Weissenborn,
  Xiaohua Zhai, Thomas Unterthiner, Mostafa Dehghani, Matthias Minderer, Georg
  Heigold, Sylvain Gelly, et~al.
\newblock An image is worth 16x16 words: Transformers for image recognition at
  scale.
\newblock In \emph{International Conference on Learning Representations}, 2020.

\bibitem[Dudley et~al.(2016)Dudley, Wickham, and
  Coombs]{dudley2016introduction}
William~N Dudley, Rita Wickham, and Nicholas Coombs.
\newblock An introduction to survival statistics: {Kaplan-Meier} analysis.
\newblock \emph{Journal of the advanced practitioner in oncology}, 7\penalty0
  (1):\penalty0 91, 2016.

\bibitem[Duong et~al.(2021)Duong, Zheng, Xia, Jin, Liu, Li, Hao, Alfreds,
  Sylvester, Widen, et~al.]{duong2021identification}
Son~Q Duong, Le~Zheng, Minjie Xia, Bo~Jin, Modi Liu, Zhen Li, Shiying Hao,
  Shaun~T Alfreds, Karl~G Sylvester, Eric Widen, et~al.
\newblock Identification of patients at risk of new onset heart failure:
  Utilizing a large statewide health information exchange to train and validate
  a risk prediction model.
\newblock \emph{Plos one}, 16\penalty0 (12):\penalty0 e0260885, 2021.

\bibitem[Eldele et~al.(2021)Eldele, Ragab, Chen, Wu, Kwoh, Li, and
  Guan]{eldele2021time}
Emadeldeen Eldele, Mohamed Ragab, Zhenghua Chen, Min Wu, Chee~Keong Kwoh,
  Xiaoli Li, and Cuntai Guan.
\newblock Time-series representation learning via temporal and contextual
  contrasting.
\newblock \emph{arXiv preprint arXiv:2106.14112}, 2021.

\bibitem[Franceschi et~al.(2019)Franceschi, Dieuleveut, and
  Jaggi]{franceschi2019unsupervised}
Jean-Yves Franceschi, Aymeric Dieuleveut, and Martin Jaggi.
\newblock Unsupervised scalable representation learning for multivariate time
  series.
\newblock \emph{Advances in neural information processing systems}, 32, 2019.

\bibitem[Gheorghiade et~al.(2005)Gheorghiade, De~Luca, Fonarow, Filippatos,
  Metra, and Francis]{Gheorghiade2005}
Mihai Gheorghiade, Leonardo De~Luca, Gregg~C. Fonarow, Gerasimos Filippatos,
  Marco Metra, and Gary~S. Francis.
\newblock Pathophysiologic targets in the early phase of acute heart failure
  syndromes.
\newblock \emph{The American Journal of Cardiology}, 96\penalty0 (6):\penalty0
  11–17, September 2005.
\newblock ISSN 0002-9149.
\newblock \doi{10.1016/j.amjcard.2005.07.016}.
\newblock URL \url{http://dx.doi.org/10.1016/j.amjcard.2005.07.016}.

\bibitem[Gopal et~al.(2021)Gopal, Han, Raghupathi, Ng, Tison, and
  Rajpurkar]{gopal20213kg}
Bryan Gopal, Ryan Han, Gautham Raghupathi, Andrew Ng, Geoff Tison, and Pranav
  Rajpurkar.
\newblock 3kg: Contrastive learning of 12-lead electrocardiograms using
  physiologically-inspired augmentations.
\newblock In \emph{Machine Learning for Health}, pages 156--167. PMLR, 2021.

\bibitem[Heiliger et~al.(2022)Heiliger, Sekuboyina, Menze, Egger, and
  Kleesiek]{heiliger2022beyond}
Lars Heiliger, Anjany Sekuboyina, Bjoern Menze, Jan Egger, and Jens Kleesiek.
\newblock Beyond medical imaging-a review of multimodal deep learning in
  radiology.
\newblock \emph{TechRxiv}, \penalty0 (19103432), 2022.

\bibitem[Hyvarinen and Morioka(2017)]{hyvarinen17a}
Aapo Hyvarinen and Hiroshi Morioka.
\newblock {Nonlinear ICA of Temporally Dependent Stationary Sources}.
\newblock In Aarti Singh and Jerry Zhu, editors, \emph{Proceedings of the 20th
  International Conference on Artificial Intelligence and Statistics},
  volume~54 of \emph{Proceedings of Machine Learning Research}, pages 460--469.
  PMLR, 20--22 Apr 2017.
\newblock URL \url{https://proceedings.mlr.press/v54/hyvarinen17a.html}.

\bibitem[Jencks et~al.(2009)Jencks, Williams, and
  Coleman]{jencks2009rehospitalizations}
Stephen~F Jencks, Mark~V Williams, and Eric~A Coleman.
\newblock Rehospitalizations among patients in the medicare fee-for-service
  program.
\newblock \emph{New England Journal of Medicine}, 360\penalty0 (14):\penalty0
  1418--1428, 2009.

\bibitem[Jeong et~al.(2023)Jeong, Stultz, and Ghassemi]{jeong2023deep}
Hyewon Jeong, Collin~M Stultz, and Marzyeh Ghassemi.
\newblock Deep metric learning for the hemodynamics inference with
  electrocardiogram signals.
\newblock In \emph{Machine Learning for Healthcare Conference}, pages 321--342.
  PMLR, 2023.

\bibitem[Johnson et~al.(2023)Johnson, Bulgarelli, Shen, Gayles, Shammout,
  Horng, Pollard, Hao, Moody, Gow, et~al.]{johnson2023mimic}
Alistair~EW Johnson, Lucas Bulgarelli, Lu~Shen, Alvin Gayles, Ayad Shammout,
  Steven Horng, Tom~J Pollard, Sicheng Hao, Benjamin Moody, Brian Gow, et~al.
\newblock {MIMIC-IV}, a freely accessible electronic health record dataset.
\newblock \emph{Scientific data}, 10\penalty0 (1):\penalty0 1, 2023.

\bibitem[King et~al.(2023)King, Yang, and Mortazavi]{pmlr-v225-king23a}
Ryan King, Tianbao Yang, and Bobak~J. Mortazavi.
\newblock Multimodal pretraining of medical time series and notes.
\newblock In Stefan Hegselmann, Antonio Parziale, Divya Shanmugam, Shengpu
  Tang, Mercy~Nyamewaa Asiedu, Serina Chang, Tom Hartvigsen, and Harvineet
  Singh, editors, \emph{Proceedings of the 3rd Machine Learning for Health
  Symposium}, volume 225 of \emph{Proceedings of Machine Learning Research},
  pages 244--255. PMLR, 10 Dec 2023.
\newblock URL \url{https://proceedings.mlr.press/v225/king23a.html}.

\bibitem[Kiyasseh et~al.(2021)Kiyasseh, Zhu, and Clifton]{kiyasseh2021clocs}
Dani Kiyasseh, Tingting Zhu, and David~A Clifton.
\newblock Clocs: Contrastive learning of cardiac signals across space, time,
  and patients.
\newblock In \emph{International Conference on Machine Learning}, pages
  5606--5615. PMLR, 2021.

\bibitem[Labach et~al.(2023)Labach, Pokhrel, Huang, Zuberi, Yi, Volkovs,
  Poutanen, and Krishnan]{labach2023DuETT}
Alex Labach, Aslesha Pokhrel, Xiao~Shi Huang, Saba Zuberi, Seung~Eun Yi,
  Maksims Volkovs, Tomi Poutanen, and Rahul~G Krishnan.
\newblock {DuETT}: Dual event time transformer for electronic health records.
\newblock In \emph{Machine Learning for Healthcare Conference}, pages 403--422.
  PMLR, 2023.

\bibitem[Lee et~al.(2023)Lee, Lee, Hahn, Hyun, Choi, Ahn, and
  Lee]{lee2023learning}
Kwanhyung Lee, Soojeong Lee, Sangchul Hahn, Heejung Hyun, Edward Choi, Byungeun
  Ahn, and Joohyung Lee.
\newblock Learning missing modal electronic health records with unified
  multi-modal data embedding and modality-aware attention.
\newblock In \emph{Machine Learning for Healthcare Conference}, pages 423--442.
  PMLR, 2023.

\bibitem[Li et~al.(2020)Li, Jamieson, Rostamizadeh, Gonina, Ben-Tzur, Hardt,
  Recht, and Talwalkar]{li2020massively}
Liam Li, Kevin Jamieson, Afshin Rostamizadeh, Ekaterina Gonina, Jonathan
  Ben-Tzur, Moritz Hardt, Benjamin Recht, and Ameet Talwalkar.
\newblock A system for massively parallel hyperparameter tuning.
\newblock \emph{Proceedings of Machine Learning and Systems}, 2:\penalty0
  230--246, 2020.

\bibitem[Li and Marlin(2020)]{li2020learning}
Steven Cheng-Xian Li and Benjamin Marlin.
\newblock Learning from irregularly-sampled time series: A missing data
  perspective.
\newblock In \emph{International Conference on Machine Learning}, pages
  5937--5946. PMLR, 2020.

\bibitem[Li et~al.(2022)Li, Mamouei, Salimi-Khorshidi, Rao, Hassaine, Canoy,
  Lukasiewicz, and Rahimi]{li2022hi}
Yikuan Li, Mohammad Mamouei, Gholamreza Salimi-Khorshidi, Shishir Rao,
  Abdelaali Hassaine, Dexter Canoy, Thomas Lukasiewicz, and Kazem Rahimi.
\newblock {Hi-BEHRT}: {Hierarchical Transformer-based} model for accurate
  prediction of clinical events using multimodal longitudinal electronic health
  records.
\newblock \emph{IEEE journal of biomedical and health informatics}, 27\penalty0
  (2):\penalty0 1106--1117, 2022.

\bibitem[Liang et~al.(2022)Liang, Zhang, Kwon, Yeung, and Zou]{liang2022mind}
Victor~Weixin Liang, Yuhui Zhang, Yongchan Kwon, Serena Yeung, and James~Y Zou.
\newblock Mind the gap: Understanding the modality gap in multi-modal
  contrastive representation learning.
\newblock \emph{Advances in Neural Information Processing Systems},
  35:\penalty0 17612--17625, 2022.

\bibitem[McDermott et~al.(2021)McDermott, Nestor, Kim, Zhang, Goldenberg,
  Szolovits, and Ghassemi]{mcdermott2021comprehensive}
Matthew B.~A. McDermott, Bret Nestor, Evan Kim, Wancong Zhang, Anna Goldenberg,
  Peter Szolovits, and Marzyeh Ghassemi.
\newblock A comprehensive {EHR} timeseries pre-training benchmark.
\newblock In \emph{Proceedings of the Conference on Health, Inference, and
  Learning}, pages 257--278, 2021.

\bibitem[McDermott et~al.(2023{\natexlab{a}})McDermott, Nestor, Argaw, and
  Kohane]{mcdermott2023event}
Matthew B.~A. McDermott, Bret Nestor, Peniel~N Argaw, and Isaac~S. Kohane.
\newblock Event stream {GPT}: A data pre-processing and modeling library for
  generative, pre-trained transformers over continuous-time sequences of
  complex events.
\newblock In \emph{Thirty-seventh Conference on Neural Information Processing
  Systems Datasets and Benchmarks Track}, 2023{\natexlab{a}}.
\newblock URL \url{https://openreview.net/forum?id=hiO0735tmc}.

\bibitem[McDermott et~al.(2023{\natexlab{b}})McDermott, Yap, Szolovits, and
  Zitnik]{mcdermott2023structure}
Matthew B.~A. McDermott, Brendan Yap, Peter Szolovits, and Marinka Zitnik.
\newblock Structure-inducing pre-training.
\newblock \emph{Nature Machine Intelligence}, pages 1--10, 2023{\natexlab{b}}.

\bibitem[McInnes et~al.(2018)McInnes, Healy, Saul, and
  Gro{\ss}berger]{mcinnes2020umap}
Leland McInnes, John Healy, Nathaniel Saul, and Lukas Gro{\ss}berger.
\newblock {UMAP: Uniform Manifold Approximation and Projection}.
\newblock \emph{Journal of Open Source Software}, 3\penalty0 (29), 2018.

\bibitem[Nguyen et~al.(2021)Nguyen, Jeong, Yang, and Hwang]{nguyen2021clinical}
A~Tuan Nguyen, Hyewon Jeong, Eunho Yang, and Sung~Ju Hwang.
\newblock Clinical risk prediction with temporal probabilistic asymmetric
  multi-task learning.
\newblock In \emph{Proceedings of the AAAI Conference on Artificial
  Intelligence}, volume~35, pages 9081--9091, 2021.

\bibitem[Oh et~al.(2022)Oh, Chung, Kwon, Hong, and Choi]{oh2022lead}
Jungwoo Oh, Hyunseung Chung, Joon-myoung Kwon, Dong-gyun Hong, and Edward Choi.
\newblock Lead-agnostic self-supervised learning for local and global
  representations of electrocardiogram.
\newblock In \emph{Conference on Health, Inference, and Learning}, pages
  338--353. PMLR, 2022.

\bibitem[Oord et~al.(2018)Oord, Li, and Vinyals]{oord2018representation}
Aaron van~den Oord, Yazhe Li, and Oriol Vinyals.
\newblock Representation learning with contrastive predictive coding.
\newblock \emph{arXiv preprint arXiv:1807.03748}, 2018.

\bibitem[Radford et~al.(2021)Radford, Kim, Hallacy, Ramesh, Goh, Agarwal,
  Sastry, Askell, Mishkin, Clark, et~al.]{radford2021learning}
Alec Radford, Jong~Wook Kim, Chris Hallacy, Aditya Ramesh, Gabriel Goh,
  Sandhini Agarwal, Girish Sastry, Amanda Askell, Pamela Mishkin, Jack Clark,
  et~al.
\newblock Learning transferable visual models from natural language
  supervision.
\newblock In \emph{International conference on machine learning}, pages
  8748--8763. PMLR, 2021.

\bibitem[Raffel and Ellis(2015)]{ffattention}
Colin Raffel and Daniel~PW Ellis.
\newblock Feed-forward networks with attention can solve some long-term memory
  problems.
\newblock \emph{arXiv preprint arXiv:1512.08756}, 2015.

\bibitem[Raghu et~al.(2023)Raghu, Chandak, Alam, Guttag, and
  Stultz]{raghu2023sequential}
Aniruddh Raghu, Payal Chandak, Ridwan Alam, John Guttag, and Collin Stultz.
\newblock Sequential multi-dimensional self-supervised learning for clinical
  time series.
\newblock In \emph{International Conference on Machine Learning}, pages
  28531--28548. PMLR, 2023.

\bibitem[Rahimi et~al.(2014)Rahimi, Bennett, Conrad, Williams, Basu, Dwight,
  Woodward, Patel, McMurray, and MacMahon]{rahimi2014risk}
Kazem Rahimi, Derrick Bennett, Nathalie Conrad, Timothy~M Williams, Joyee Basu,
  Jeremy Dwight, Mark Woodward, Anushka Patel, John McMurray, and Stephen
  MacMahon.
\newblock Risk prediction in patients with heart failure: a systematic review
  and analysis.
\newblock \emph{JACC: Heart Failure}, 2\penalty0 (5):\penalty0 440--446, 2014.

\bibitem[Rajkomar et~al.(2018)Rajkomar, Oren, Chen, Dai, Hajaj, Hardt, Liu,
  Liu, Marcus, Sun, et~al.]{rajkomar2018scalable}
Alvin Rajkomar, Eyal Oren, Kai Chen, Andrew~M Dai, Nissan Hajaj, Michaela
  Hardt, Peter~J Liu, Xiaobing Liu, Jake Marcus, Mimi Sun, et~al.
\newblock Scalable and accurate deep learning with electronic health records.
\newblock \emph{NPJ digital medicine}, 1\penalty0 (1):\penalty0 18, 2018.

\bibitem[Satopaa et~al.(2011)Satopaa, Albrecht, Irwin, and Raghavan]{kneedle}
Ville Satopaa, Jeannie Albrecht, David Irwin, and Barath Raghavan.
\newblock Finding a "kneedle" in a haystack: Detecting knee points in system
  behavior.
\newblock In \emph{2011 31st International Conference on Distributed Computing
  Systems Workshops}, pages 166--171, 2011.
\newblock \doi{10.1109/ICDCSW.2011.20}.

\bibitem[Shukla and Marlin(2020)]{shukla2020survey}
Satya~Narayan Shukla and Benjamin~M Marlin.
\newblock A survey on principles, models and methods for learning from
  irregularly sampled time series.
\newblock \emph{arXiv preprint arXiv:2012.00168}, 2020.

\bibitem[Shwartz-Ziv and Armon(2022)]{shwartzziv2021tabular}
Ravid Shwartz-Ziv and Amitai Armon.
\newblock Tabular data: Deep learning is not all you need.
\newblock \emph{Information Fusion}, 81:\penalty0 84--90, 2022.

\bibitem[Tipirneni and Reddy(2022)]{tipirneni2022self}
Sindhu Tipirneni and Chandan~K Reddy.
\newblock Self-supervised transformer for sparse and irregularly sampled
  multivariate clinical time-series.
\newblock \emph{ACM Transactions on Knowledge Discovery from Data (TKDD)},
  16\penalty0 (6):\penalty0 1--17, 2022.

\bibitem[Tonekaboni et~al.(2021)Tonekaboni, Eytan, and
  Goldenberg]{tonekaboni2021unsupervised}
Sana Tonekaboni, Danny Eytan, and Anna Goldenberg.
\newblock Unsupervised representation learning for time series with temporal
  neighborhood coding.
\newblock \emph{arXiv preprint arXiv:2106.00750}, 2021.

\bibitem[Upadhyay et~al.(2019)Upadhyay, Stephenson, and
  Smith]{upadhyay2019readmission}
Soumya Upadhyay, Amber~L Stephenson, and Dean~G Smith.
\newblock Readmission rates and their impact on hospital financial performance:
  a study of washington hospitals.
\newblock \emph{INQUIRY: The Journal of Health Care Organization, Provision,
  and Financing}, 56:\penalty0 0046958019860386, 2019.

\bibitem[Wang et~al.(2020)Wang, McDermott, Chauhan, Ghassemi, Hughes, and
  Naumann]{wang2020mimic}
Shirly Wang, Matthew~BA McDermott, Geeticka Chauhan, Marzyeh Ghassemi,
  Michael~C Hughes, and Tristan Naumann.
\newblock Mimic-extract: A data extraction, preprocessing, and representation
  pipeline for mimic-iii.
\newblock In \emph{Proceedings of the ACM conference on health, inference, and
  learning}, pages 222--235, 2020.

\bibitem[Zhang and Kuo(2024)]{Zhang2024}
Min Zhang and Tsung-Ting Kuo.
\newblock Early prediction of long hospital stay for intensive care units
  readmission patients using medication information.
\newblock \emph{Computers in Biology and Medicine}, 174:\penalty0 108451, May
  2024.
\newblock ISSN 0010-4825.
\newblock \doi{10.1016/j.compbiomed.2024.108451}.
\newblock URL \url{http://dx.doi.org/10.1016/j.compbiomed.2024.108451}.

\bibitem[Zhang et~al.(2021)Zhang, Zeman, Tsiligkaridis, and
  Zitnik]{zhang2021graph}
Xiang Zhang, Marko Zeman, Theodoros Tsiligkaridis, and Marinka Zitnik.
\newblock Graph-guided network for irregularly sampled multivariate time
  series.
\newblock In \emph{International Conference on Learning Representations}, 2021.

\bibitem[Zhang et~al.(2022)Zhang, Jiang, Miura, Manning, and
  Langlotz]{zhang2022contrastive}
Yuhao Zhang, Hang Jiang, Yasuhide Miura, Christopher~D Manning, and Curtis~P
  Langlotz.
\newblock Contrastive learning of medical visual representations from paired
  images and text.
\newblock In \emph{Machine Learning for Healthcare Conference}, pages 2--25.
  PMLR, 2022.

\end{thebibliography}

\newpage
\appendix
\onecolumn

\section{Views of Contrastive Learning}

In this section, we outline the contrastive learning framework of EBCL and its variations compared to the contrastive baseline (OCP \citep{agrawal2022leveraging}) and specify some examples of what information they learn throughout pretraining. Our proposed method (EBCL) takes positive and negative samples (A, B) or (C, D) from the same patient’s medical record, where window A or C immediately precedes window B or D in time in the original medical record (Algorithm \ref{clip}). Furthermore, windows A and C immediately precede a key medical event (e.g., inpatient admission), and windows B and D immediately follow that same key medical event in the original medical record. EBCL uses negative sample pairs (A, D) or (C, B) from different patients’ medical records, where, for instance, A immediately precedes a key medical event and D immediately follows a key medical event in the original medical record. Our EBCL variants (EBCL Censored, EBCL Outpatients) follow the same definition for selecting positive and negative pairs, where we censor some windows around the index event (EBCL Censored) or select different index event (EBCL Outpatients). On the contrary, OCP takes the swapped sequence (N, M) from the same patient to become the negative sample and the original ordered sequence (M, N) becomes the positive sample.

\begin{figure*}[h]
\centering
  \includegraphics[width=0.8\textwidth]{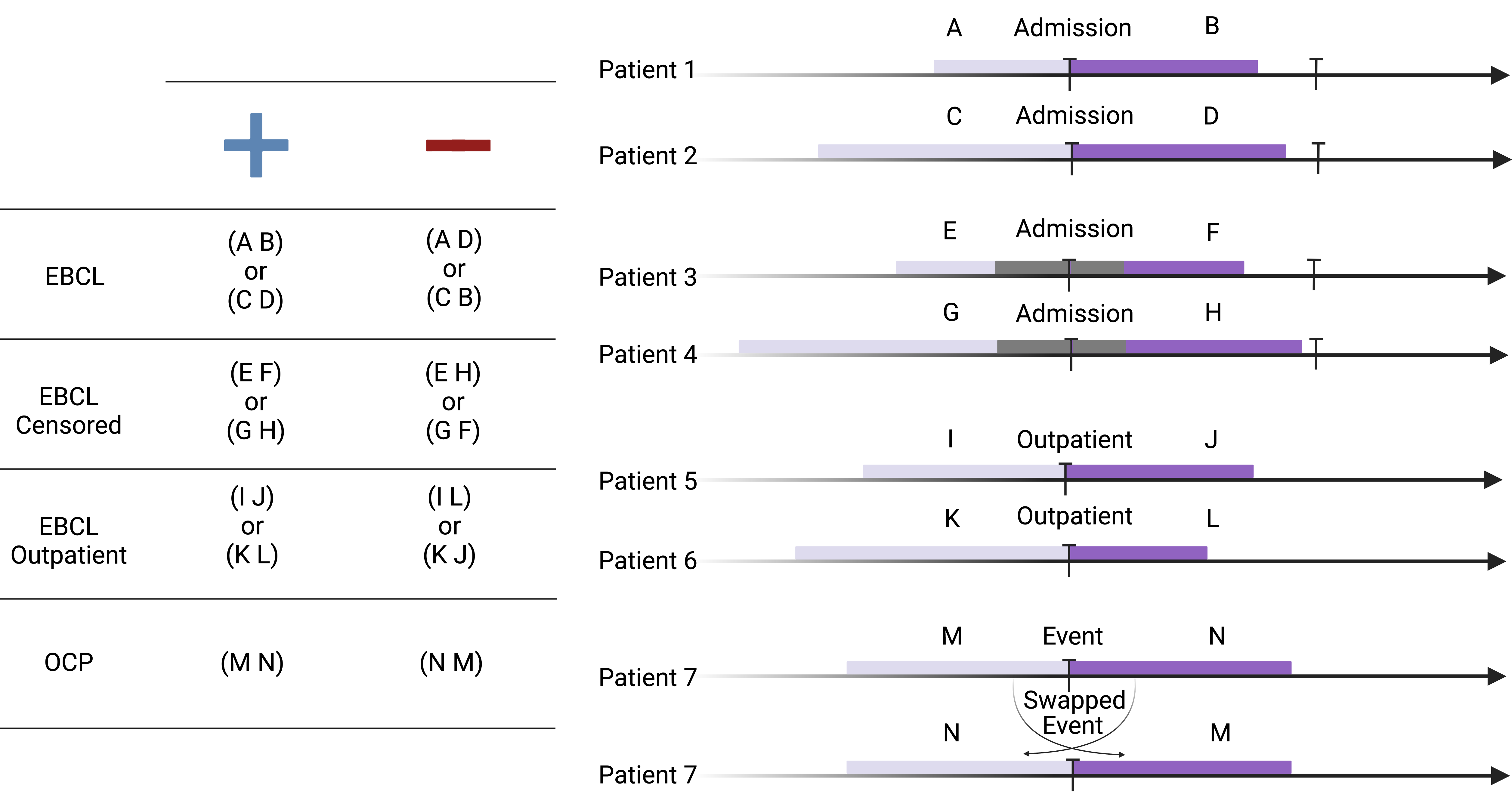}
  \caption{\textbf{Contrastive Learning Framework of EBCL, EBCL Ablations (EBCL Censored, EBCL Outpatient) and OCP.}}
  \label{fig:positive}
\end{figure*}
\begin{algorithm}
\caption{Event-Based Contrastive Learning}\label{clip}
\KwIn{$(\text{pre}_i, \text{post}_i)_{i=1}^{N}$ pre, post time-series patient dataset pairs}
\For{$\text{epoch} = 1$ \KwTo $M$}{
    \For{$\text{batch\_id} = 1$ \KwTo $K$}{
        Sample minibatch $(PRE, POST)= (\text{pre}_i, \text{post}_i)_{i=1}^{N/K}$ \tcp*{encode pre and post trajectories}
        $PRE_f \gets \text{transformer\_encoder}(PRE)$\\
        $POST_f \gets \text{transformer\_encoder}(POST)$ \tcp*{event based contrastive learning}
        $PRE_e \gets \text{l2\_normalize}(\text{np.dot}(PRE_f, W_{pre}), \text{axis}=1)$\\
        $POST_e \gets \text{l2\_normalize}(\text{np.dot}(POST_f, W_{post}), \text{axis}=1)$ \tcp*{scaled pairwise cosine similarities}
        $\text{logits} \gets \text{np.dot}(PRE_e, POST_e^T) \times \text{np.exp}(t)$ \tcp*{symmetric loss function}
        $\text{labels} \gets \text{np.arange}(n)$\\
        $\text{loss\_i} \gets \text{CE}(\text{logits}, \text{labels}, \text{axis}=0)$\\
        $\text{loss\_t} \gets \text{CE}(\text{logits}, \text{labels}, \text{axis}=1)$\\
        $\text{loss} \gets (\text{loss\_i} + \text{loss\_t})/2$
    }
}
\end{algorithm}


Due to this difference inherently arising from the definition of positive and negative samples, OCP \citep{agrawal2022leveraging} and EBCL have distinct potential for learning different types of features in a medical record. We can think of two different features that have $x_1$ a static nature that doesn't change frequently over time and $x_2$ a time-varying feature that changes over a short period of time. Under OCP pre-training, the value of feature $x_1$ is identical over two consecutive windows M and N. Thus, OCP pre-training cannot use the observed value of the feature to differentiate between a positive pair (M, N) and a negative pair (N, M) as both pairs agree on the feature value, so the model is not incentivized to capture the approximate value of the static feature in the produced embeddings at all. For EBCL, the value of the static feature agrees between positive samples, as it originates from the same patient. However, the value of the feature will disagree in the negative sample. Therefore, the EBCL model is incentivized to capture the approximate value of static features in its embeddings.

Let's now think about a feature that oscillates over time, which stays centered near a constant value, but oscillates with local temporal trends across both patients. Under OCP pretraining, the model can see that the trends observed in window M will directly continue into window N, which gives a strong signal that (M, N) is an appropriately ordered pair. In contrast, for a swapped sequence, the model will see that the trends do not continue from N to M, which clearly indicates an incorrectly ordered pair. Thus, this will be a strong signal for the model to differentiate positive (correctly ordered) from negative (incorrectly ordered) pairs, so the model will be incentivized to capture the local trends of oscillating features within windows M and N. Under EBCL pretraining, much like in OCP, the model can see that the trends observed in window A will directly continue into window B, which gives a strong signal that (A, B) is an appropriately ordered pair within the same patient. In contrast, for the negative pair (A, D) the model will be much less likely to observe a non-continuing pattern, which will be a clear negative signal. So, similar to OCP, the model is incentivized to capture the local trends of features within windows A and B in its embeddings. This is under the assumption that the feature is likely to be observed and retain its locally oscillatory behavior both near key events and distant from key events.

\section{Dataset and Experimental Setting}
\label{apd:dataset}

\begin{figure*}[t]
\centering
  \includegraphics[width=0.8\textwidth]{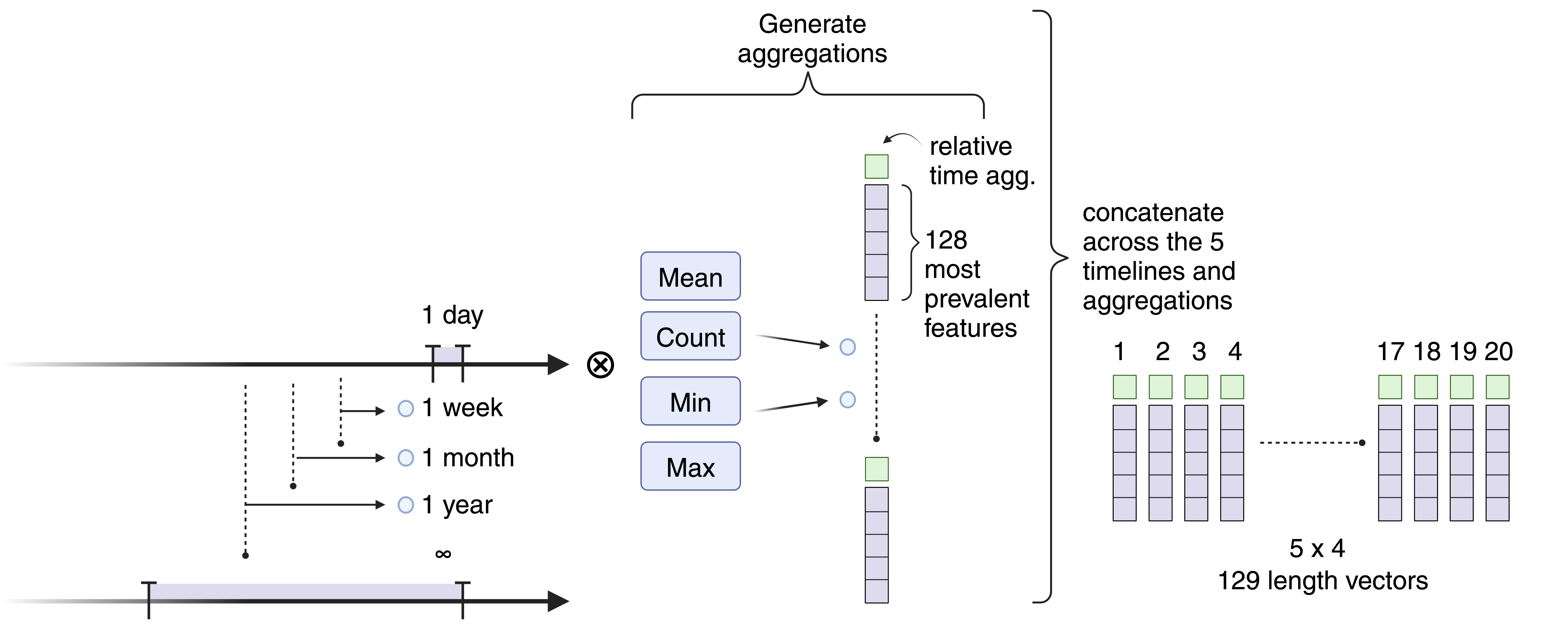}
  \caption{\textbf{XGBoost} requires a fixed size vector as input. We create this vector from time series data as follows: we restrict to the 128 most prevalent features in the data and the relative time of those observations (relative to the time of the decision). For each of the time windows prior to the time of the decision. For the heart failure cohort, windows end at discharge time for 1-Year Mortality and 30-Day Readmission and admission time for 7-Day LOS. For the MIMIC-IV cohort, windows end at the event time (i.e. hypotension or mechanical ventillation onset time). We perform feature aggregations (mean, count, min value, max value) for each feature over all pre-defined time windows. We then concatenate all of the feature aggregation outputs across all windows to get a final vector that is the input to XGBoost. We have 5 window sizes and 4 aggregations each generating a 129 length vector. After concatenating these, we get a 2,580 size vector that is the input for XGBoost.} \label{fig:xgboost}
\end{figure*}
\begin{figure*}[t]
\centering
  \includegraphics[width=0.8\textwidth]{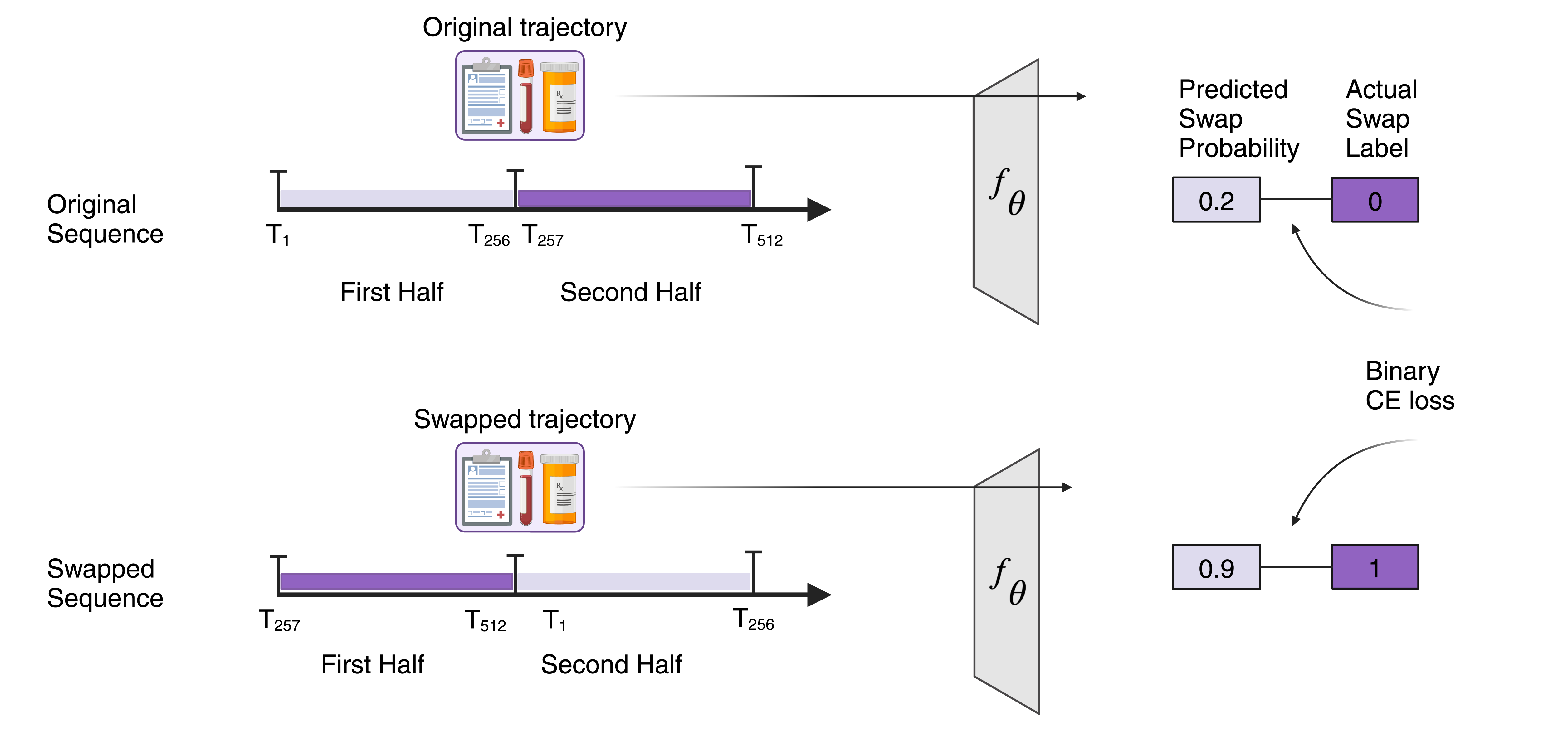}
  \caption{\textbf{OCP Positive and Negative Pairs}. In OCP, positive pairs are sequences with correct order, while negative pairs have the order of the first and second halves of the timeline swapped. The model, $f_\theta$ is pretrained to predict whether the sequence was swapped.}
  \label{fig:ocp}
\end{figure*}
\begin{figure*}[t]
\centering
  \includegraphics[width=0.8\textwidth]{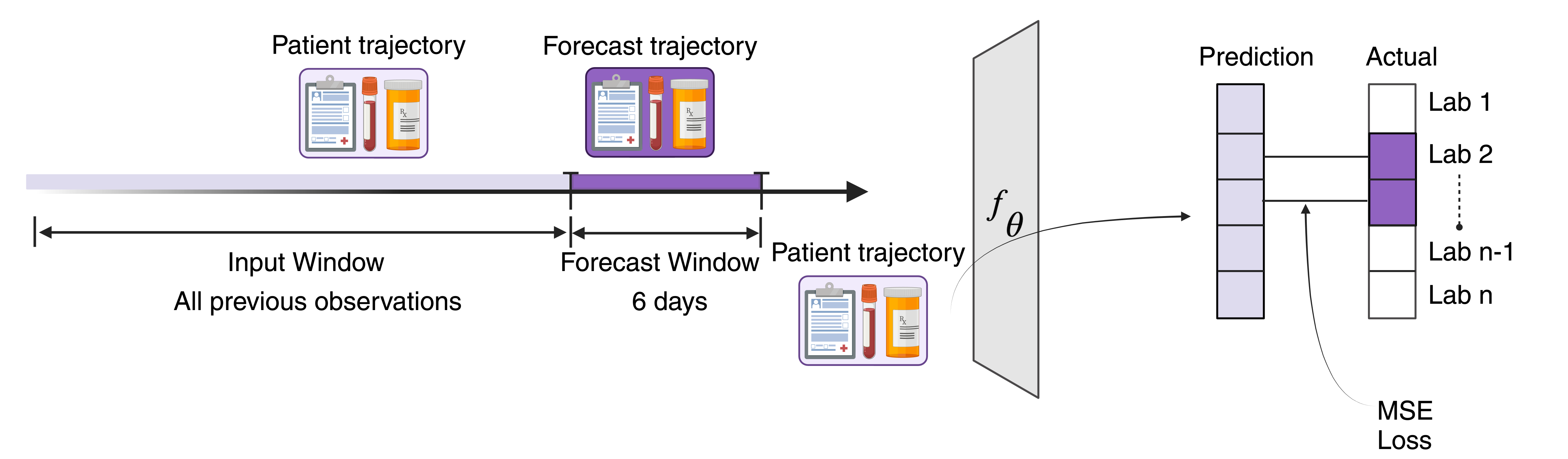}
  \caption{\textbf{STraTS} forecast patient trajectory in the forecasting window and is trained with Mean Squared Error (MSE) loss calculated on the observations present in the forecasting window.}
  \label{fig:strats}
\end{figure*}
\begin{figure*}[t]
\centering
  \includegraphics[width=0.8\textwidth]{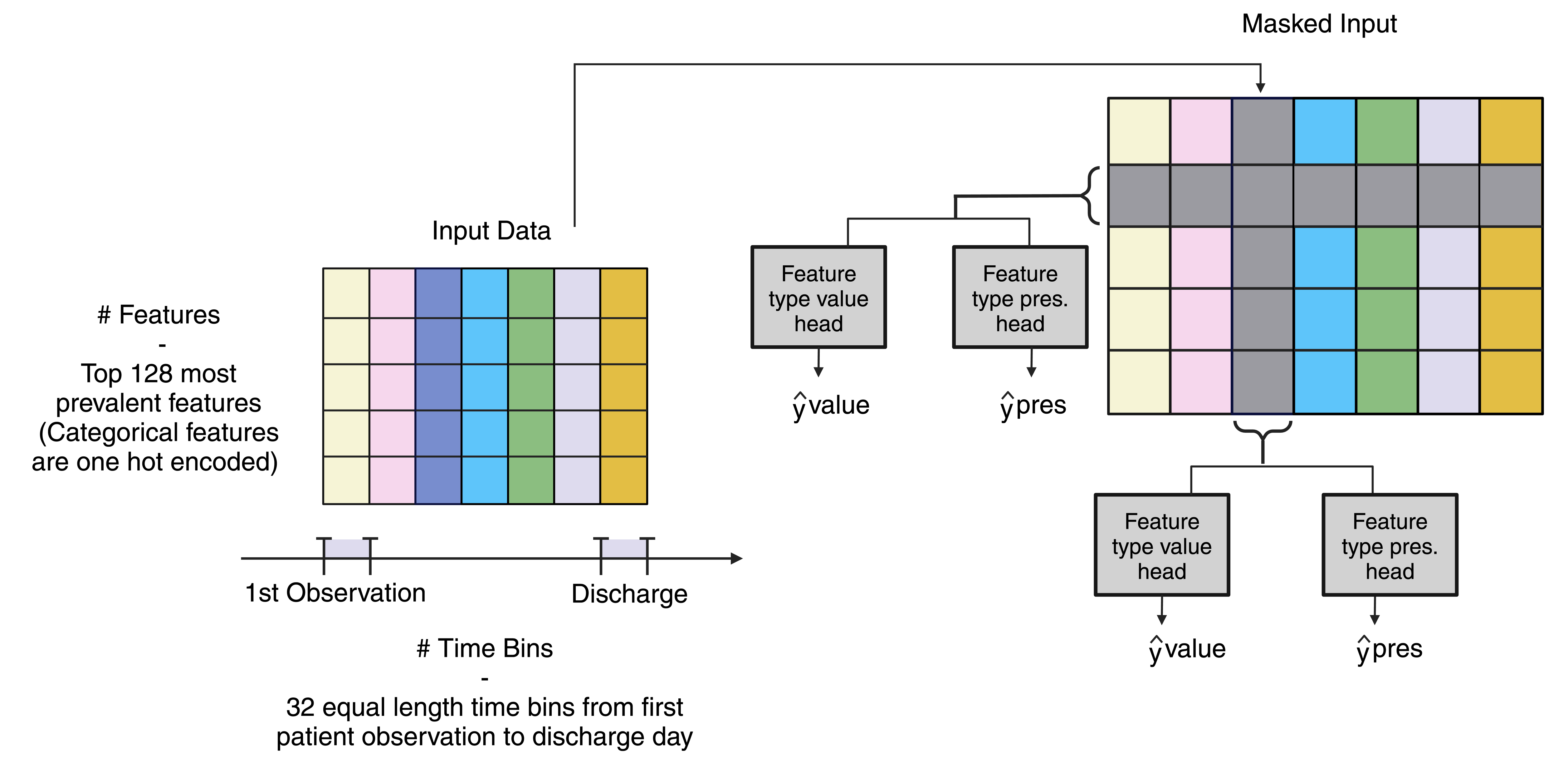}
  \caption{\textbf{DueTT} bins data into 32 time bins and a finite number of events as is depicted on the left side. For our admissions data this corresponds to binning data from the patient's first observation to their discharge time (or admission time for the LOS task). The pretraining task is to mask out a row (an event) and a column (a time bin) and predict the event values and event presences for the masked blocks.}
  \label{fig:duett}
\end{figure*}

\subsection{Dataset Preprocessing}
We preprocess our heart failure dataset as follows: features with less than $1,000$ occurrences in the entire dataset are dropped. Categorical values with less than $1,000$ occurrences are replaced with the categorical value ``UNKNOWN''. We use the triplet embedding strategy from \citep{tipirneni2022self} for modeling sequential EHR data, and this allows flexibility in how dates are encoded. For all experiments, we encode dates as the relative time in days from the inpatient admission event divided by the standard deviation of these times in the training set. We label encode categorical observation values and features, and z normalize continuous data and feed it into a continuous value embedder to get a vector.




\subsection{Task Definition}
We pretrained the transformer on the time-series data around the domain knowledge-driven, cohort-specific important events for each patient. We have three pretrained models where each model was trained on one of three different cohort/event combinations: heart failure cohort on admission event, MIMIC-IV ICU cohort on hypotension event, and MIMIC-IV ICU cohort on mechanical ventilation. Each model pretrained around the important event of each cohort is finetuned to predict the outcome for the same patient. 

\begin{table}[h!]
\centering
\caption{Task Definition}
\label{table:dataset}
\begin{tabular}{l l l}
\toprule
\textbf{Dataset} & \textbf{Cohort/Event} & \textbf{Fine-tuning Task} \\
\midrule
Heart Failure & Inpatient Admission & Mortality/LOS/Readmission \\ 
MIMIC-IV & Hypotension & Mortality/LOS \\ 
MIMIC-IV & Mechanical Ventilation & Mortality/LOS \\ 
\bottomrule
\end{tabular}
\end{table}

\subsection{Model Architecture and Training}
\begin{figure*}[htbp]
\centering
  \includegraphics[width=.6\textwidth]{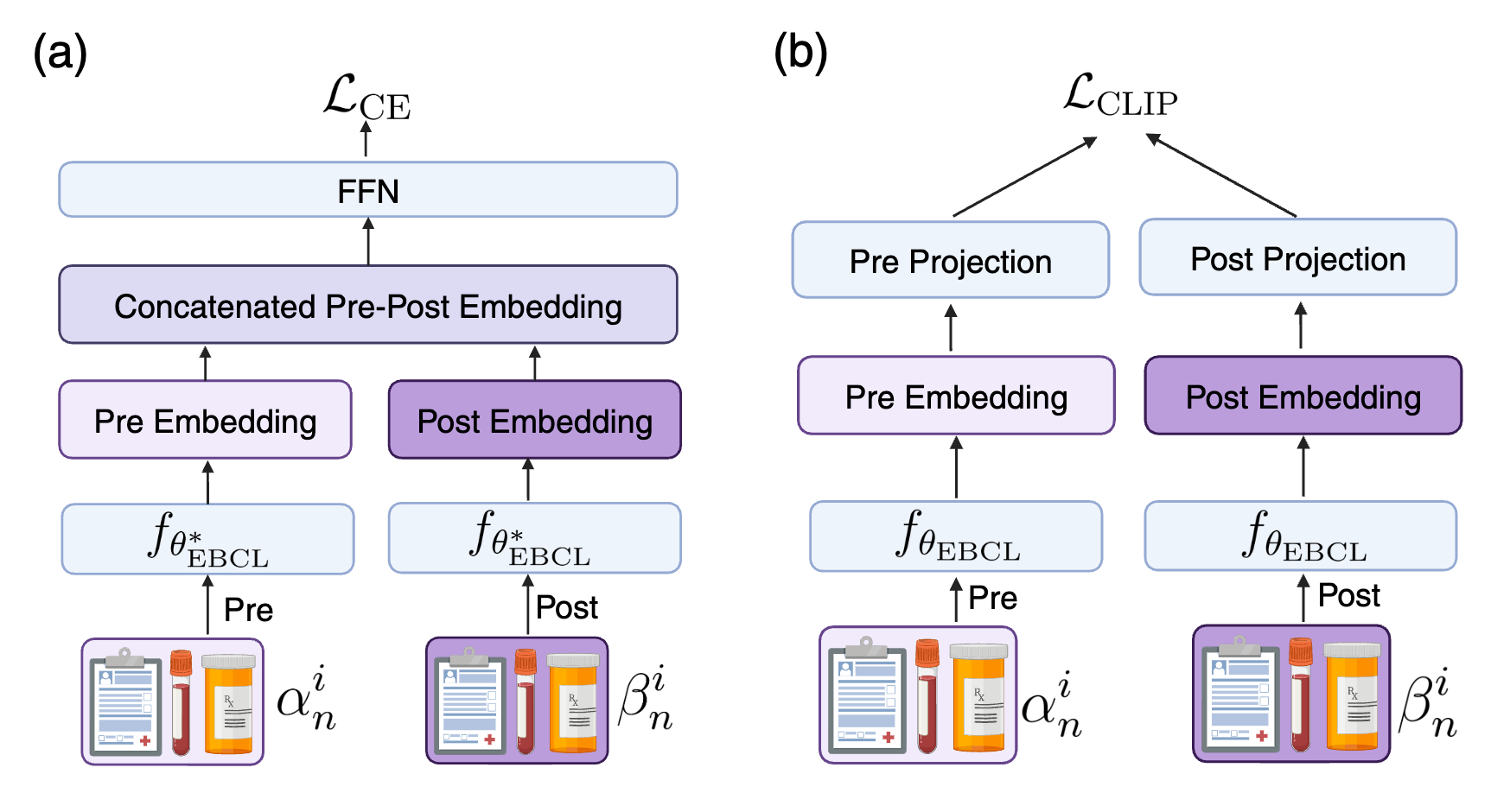}
  \caption{\textbf{Model Architectures}. FFN refers to a one hidden layer feed-forward network. \textbf{(a)} This architecture is used for EBCL, OCP, STraTS, and Supervised experiments. For EBCL we initialize $\theta$ with the final weights from EBCL pretraining, $\theta^*_\mathrm{EBCL}$, for OCP we initialize with the final weights from OCP pretraining, $\theta^*_\mathrm{OCP}$, and for Supervised we randomly initialize $\theta$. Notice that Pre-event data and Post-event data are passed separately into the transformer encoder $f_\theta$, so there is no self-attention between Pre-event and Post-event data. For Pre-event-only experiments, such as predicting 7-day LOS for the heart failure cohort, we leave out the 'Post Embedding' part of this architecture. \textbf{(b)} Architecture used for OCP Pretraining. The binary label for computing the cross-entropy loss, $\mathcal{L}\textsubscript{CE}$, is whether or not the random sequence, $\tau^i$, is swapped. \textbf{(c)} Architecture used for EBCL Pretraining. The "Pre Projection" and "Post Projection" are linear projections, and their weights are not shared.}
  \label{fig:architecture}
\end{figure*}

Figure \ref{fig:architecture} summarizes the model architecture used for pretraining and finetuning for EBCL, OCP, STraTS, and fully supervised experiments. 

\paragraph{Tokenization Method} We model patient trajectory data as a sequence of observations, represented as triples consisting of time (e.g., the time of observation), feature (e.g., a creatinine lab test), and value (e.g., the creatine lab test result) inspired by STraTS \cite{tipirneni2022self}. We employ different encoders for each element of the triplet: a one-to-many feed-forward network for time, a one-to-many feed-forward network for continuous values, and a lookup table for embedding categorical features and values. These embeddings are then summed together to create the tokenized input for our model. 

\paragraph{Hyperparameter tuning} \label{apd:tune} For both our pretraining and finetuning, we perform the same hyperparameter search with $16$ pairs of randomly sampled learning rates and dropouts. Learning rates are sampled from the log uniform distribution ($1e-6$, $1e-2$), and dropouts are sampled from the uniform distribution ($0$, $0.6$). We use the ASHA scheduler \citep{li2020massively} with a grace period of $4$ epochs and a reduction factor of $2$ to schedule the training of these jobs and select the trial with the best final validation loss. The ASHA scheduler will halt trials early in training that have a very high loss relative to other trials. Additionally, we use an early stopping tolerance of $3$ epochs for all experiments except for OCP and DuETT which have more unstable training dynamics, in which case we use a tolerance of $10$.

\subsection{XGBoost} 
For the preparation of input dataset for XGBoost, we selected the 128 most prevalent features in our dataset and added the 129th feature for the relative time of these observations from admissions. We aggregate summary statistics (average, minimum, maximum, count) of the dataset within different windows (1 day, 1 week, 1 month, 1 year, $\infty$ years) from the decision date (Figure \ref{fig:xgboost}). For a fair comparison, we limit XGBoost to the same observations available in the EBCL pre and post windows (Table \ref{tab:pred}).


\subsection{Order Contrastive Pretraining \citep{agrawal2022leveraging}} We prepare a continuous trajectory of 512 consecutive data points. This trajectory might either be maintained in its original order or have its two halves swapped (Figure \ref{fig:ocp}. For the case where we keep the original ordering, the last data point, $T_{256}$, is set to time $0$, with subsequent data points indicating the time elapsed since $T_{256}$. Alternatively, if we swap the trajectory,  the last data point of the latter half, $T_{512}$, becomes time $0$. Other data points then denote the time difference from $T_{512}$. We further adjust the dates of the initial half by adding the time gap, $T\textsubscript{GAP} = T_{512} - T_{256}$. This adjustment ensures that the gap between $T_{256}$ and $T_{257}$ remains unaltered, regardless of whether the sequence is swapped or not. 

For OCP pretraining we use the same transformer model as the EBCL experiments use, and for finetuning the same EBCL finetuning architecture displayed in figure \ref{fig:architecture} (a). Due to slow convergence, we allow up to a maximum of 300 epochs for pretraining. For fine-tuning, we load the OCP pretrained weights into our finetuning architecture in Figure \ref{fig:architecture}.

\subsection{STraTS \citep{tipirneni2022self}}
STraTS represents time series as observation triplets, utilizing Continuous Value Embedding for time, feature, and values, and incorporates self-supervised learning for better generalization in data-limited scenarios. We implement the STraTS forecasting strategy for our dataset by randomly sampling a forecasting window with at least one observation in it and 16 observations prior to it. For the heart failure cohort, we use the median inpatient length of stay, 6 days, as our forecasting window length. For the MIMIC-IV ICU cohort, we use 2-hours just as STraTS did in their ICU experiments. All observations prior to the forecasting window (cutoff at a maximum of 512) are in our input window, and the STraTS pretraining task is to predict the data values in the prediction window. We calculate the Mean Squared Error (MSE) loss between features present in the forecast window with the forecasted output, as shown in Figure \ref{fig:strats}. 

The input window is all data randomly sampled before the forecast window of 6 days, where we have at least 1 observation in the forecast window and 16 in the input window. All observations before the forecasting window (cutoff at a maximum of 512) are in our input window, and the pretraining task is to predict the data values in the prediction window, and the loss is computed over the subset of values that are observed in the prediction window (Figure \ref{fig:strats}). We use the same relative times as described for the original sequence in the OCP experiment. For STraTS pretraining we use the same transformer model as the EBCL experiments use, and for finetuning the same EBCL finetuning architecture (Figure \ref{fig:architecture} (a)).

\subsection{DuETT \citep{labach2023DuETT}}
We select the $128$ most prevalent features in the dataset and bin the dataset into 32-time bins. We augment this two-dimensional matrix of input data with 128 features and 32-time bins by stacking the same size of the input matrix with the count of each feature within each time bin. We use two transformer encoder layers over the time dimension and two over the feature dimensions, as performed in the DuETT paper. See Appendix Figure \ref{fig:duett} for a visual. We generally achieved the best performance pretraining on timebins covering pre and post-data combined, even when we finetuned on only pre or post-data, such as for the LOS task. For producing embeddings, we use a DuETT model pretrained on time bins covering pre and post-data. We generate a pre-embedding by inputting only pre event data and averaging the output tensor over the $128$ feature dimensions and $32$ time bin dimensions to get a single $24$ dimensional pre data vector. We get a $24$-dimensional post embedding the same way.

\section{KNN}\label{app:knn}
For our KNN Classifier experiments, we do a sweep over all combinations of the following parameters:
\begin{enumerate}
    \item \textbf{Neighbor Weighting}: uniform or distance
    \item \textbf{KNN Model}: Pre-and-Post or ensemble
    \item \textbf{Distance Metrics} Cosine distance, Euclidean, or Euclidean with pre and post embeddings individually L2 normalized.
    \item \textbf{Number of Neighbors} 10, 30, 100, 300, and 1000
\end{enumerate}

\begin{figure*}[t]
\centering
  \includegraphics[width=\textwidth]{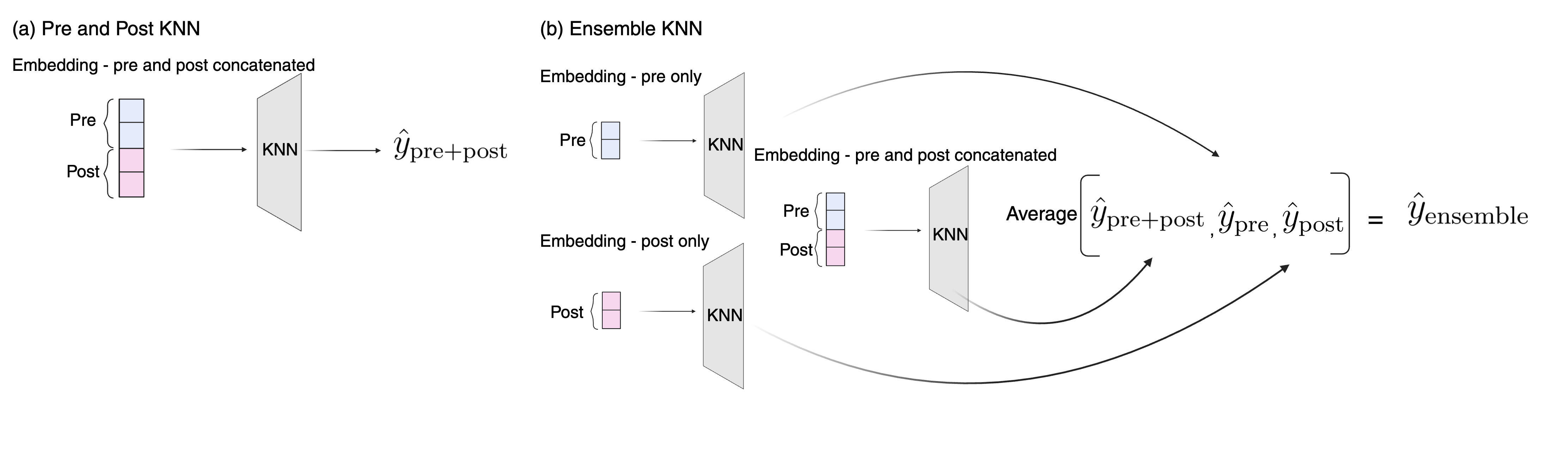}
  \caption{\textbf{KNN Models.} We try two KNN classifier strategies. The \textbf{Pre and Post} model concatenates the pre and post window representations before feeding them into the KNN. The \textbf{Ensemble} model averages the predicted class probabilities from three KNNs where: (1) $\hat{y}\textsubscript{pre}=$ class probabilities from a KNN retrieving pre representation neighbors, (2) $\hat{y}\textsubscript{post}=$ class probabilities from a KNN retrieving post representation neighbors, or (3) $\hat{y}\textsubscript{pre+post}=$ class probabilities from a KNN retrieving neighbors of the concatenated pre and post representation.
  }
  \label{fig:knn}
\end{figure*}
\begin{table*}[h]
\caption{\textbf{K-Nearest Neighbors for Evaluating the Pretrained Embeddings.} We boldface the best performing model.}
\footnotesize
\label{tab:knn_all}
\begin{center}
  \begin{tabular}{lcccccc}
  \toprule
    & \multicolumn{2}{c}{30-Day Readmission} &\multicolumn{2}{c}{1-Year Mortality} &\multicolumn{2}{c}{1-Week LOS} \\
    \cmidrule(r){2-3}\cmidrule(r){4-5}\cmidrule(r){6-7}
    & AUC & APR  & AUC & APR & AUC & APR \\
    \midrule
    OCP 
    & 67.46 $\pm$ 0.36 
    & 84.00 $\pm$ 0.30 
    & 72.61 $\pm$ 0.39 & 84.68 $\pm$ 0.35 & 64.51 $\pm$ 0.29 & 61.90 $\pm$ 0.39 
    \\
    STraTS 
    & 60.49 $\pm$ 0.37 & 79.95 $\pm$ 0.34 & 64.06 $\pm$ 0.43 & 78.77 $\pm$ 0.41 & 57.84 $\pm$ 0.30 & 53.57 $\pm$ 0.41 
    \\
    DuETT 
    & 59.71 $\pm$ 0.38 & 79.32 $\pm$ 0.34 &  64.32 $\pm$  0.42 & 78.06 $\pm$ 0.43 & 57.17 $\pm$ 0.30	& 53.16 $\pm$ 0.40 
    \\
    \midrule
    EBCL 
    & \textbf{68.46 $\pm$ 0.36} & \textbf{84.89 $\pm$ 0.28} & \textbf{77.87 $\pm$ 0.34} & \textbf{88.21 $\pm$ 0.28} & \textbf{81.59 $\pm$ 0.22} & \textbf{81.66 $\pm$ 0.25} \\
\bottomrule
\end{tabular}
\end{center}
\end{table*}

For neighbor weighting, we either weigh the labels of neighbors uniformly or by the inverse of their distance. We detail KNN models in Figure \ref{fig:knn}. The tasks that use pre and post-representations (1-Year Mortality and 30-Day Readmission) use either of two KNN classifier models. For the \texttt{Pre and Post} KNN model, the pre and post-representations are concatenated, and we fit a KNN classifier on these concatenated embeddings. For the \texttt{Ensemble} KNN model, we average the class probabilities from a \texttt{Pre and Post}, a pre-only, and a post-only KNN classifier. Note that for the 1-Week LOS task, we use only a pre-only KNN model. For each model's embeddings, we select the hyperparameter combination with the best validation set performance and report the test set performance of this configuration. To obtain standard deviations for K-NN results we use bootstrapping: we sample with replacement 1000 bootstraps from the testing dataset, and report mean and standard deviation of each metric computed with these samples.
We achieve similar results to linear probing and again observe that EBCL significantly outperforms other baselines. Moreover, this analysis reveals that neighbors within the EBCL latent space are more similar in outcomes than those derived from any other baseline model, suggesting that EBCL inherently learns an outcome-related clustering structure through pretraining alone.

\section{Results}
We have analyzed the area under the precision-recall curve (AUPRC) across three different events and two datasets—HF dataset and MIMIC-IV ICU—applied to three distinct outcome prediction tasks. Consistent with the previously summarized AUROC results, EBCL outperformed other baseline methods when applied to the HF dataset (Table \ref{tab:pred_apr}). It also excelled in three out of four event/task combinations for the MIMIC-IV ICU dataset (Table \ref{tab:mimic_pred_apr}). Additionally, the linear probing results align with those observed in the AUROC assessments (Table \ref{tab:linear_probe_apr}, \ref{tab:mimic_linear_probe_apr}), demonstrating that the frozen weights of the EBCL model predict outcomes more accurately than the frozen embeddings from other baseline methodologies. This consistency reinforces EBCL's robustness and effectiveness in clinical outcome prediction across multiple settings.

\begin{table*}[h!]
\caption{\textbf{EBCL Pretraining improves results over a supervised baseline and time-series pretraining baselines in the Heart Failure Dataset.} For the heart failure cohort, we summarize the downstream finetuning performance using both the area under precision recall curve (AUPRC) of three prediction tasks (30-Day Readmission, 1-Year Mortality, and 1-Week Length of Stay (LOS)) averaged over 5 runs with different seeds. We present finetuning performance on the MIMIC-IV Dataset for the outcomes of in-ICU Mortality and 3-Day LOS. Across all tasks, EBCL results (boldfaced) were statistically significantly better than all other tested models.
}
\label{tab:pred_apr}
\small
\begin{center}
  \begin{tabular}{lccccccc}
  \toprule
    & \multicolumn{3}{c}{Heart Failure Cohort} \\
    \cmidrule(r){2-4}
    & 30-Day Readmission
    & 1-Year Mortality
    & 1-Week LOS 
    \\
    \midrule
    \midrule
    XGBoost
    & 86.19 $\pm$ 0.03
    & 89.58 $\pm$ 0.08
    & 79.79 $\pm$ 0.08
    \\
    S-Trans
    & 85.60 $\pm$ 0.12
    & 90.27 $\pm$ 0.15
    & 88.61 $\pm$ 0.48
    \\
    \midrule
    OCP 
    & 85.48 $\pm$ 0.20
    & 89.36 $\pm$ 0.18
    & 90.05 $\pm$ 0.28
    \\
    STraTS 
    & 85.45 $\pm$ 0.07
    & 89.31 $\pm$ 0.43
    & 87.73 $\pm$ 2.03
    \\
    DueTT 
    & 85.51 $\pm$ 0.31
    & 89.04 $\pm$ 0.11
    & 74.66 $\pm$ 0.62 
    \\
    \midrule
    EBCL
    & \textbf{86.40 $\pm$ 0.03}
    & \textbf{90.55 $\pm$ 0.03}
    & \textbf{90.96 $\pm$ 0.04}
    \\
    \bottomrule
\end{tabular}
\end{center}
\end{table*}

\begin{table*}[h!]
\caption{\textbf{Linear Probing for Evaluating the Pretrained Embeddings.} EBCL (boldfaced) yielded the embedding that performs the best across all tasks.}
\small
\label{tab:linear_probe_apr}
\begin{center}
  \begin{tabular}{lccc}
  \toprule
    & 30-Day Readmission
    & 1-Year Mortality
    & 1-Week LOS
    \\
    \midrule
    OCP 
& 82.50 ± 0.27 & 83.05 ± 0.56 & 55.08 ± 0.88
    \\
    STraTS 
& 79.94 ± 0.43 & 79.34 ± 0.89 & 53.34 ± 0.94
    \\ 
    DuETT 
& 78.44 ± 2.52 & 78.04 ± 4.82 & 50.32 ± 1.22
    \\ 
    \midrule
    EBCL
& \textbf{82.92 ± 0.06} & \textbf{85.85 ± 0.23} & \textbf{69.67 ± 3.26}
    \\
\bottomrule
\end{tabular}
\end{center}
\end{table*}
\begin{table*}[h!]
\caption{\textbf{EBCL Pretraining improves downstream outcome prediction over a supervised baseline and time-series pretraining baselines in the MIMIC-IV cohort.} We summarize the finetuning performance using two metrics: area under the precision-recall curve (APR) of two prediction tasks (In-ICU Mortality and 3-Days Length of Stay (LOS)). The result has averaged over 5 runs with different seeds.
}
\label{tab:mimic_pred_apr}
\small
\begin{center}
  \begin{tabular}{lccccccc}
  \toprule
    &\multicolumn{2}{c}{MIMIC-IV Hypotension} &\multicolumn{2}{c}{MIMIC-IV Mechanical Ventilation}\\
    \cmidrule(r){2-3}\cmidrule(r){4-5}
    & In-ICU Mortality & 3-Day LOS & In-ICU Mortality & 3-Day LOS \\
    \midrule
    \midrule
    XGBoost
    & \textbf{55.34 $\pm$ 0.62}
    & 55.64 $\pm$ 2.72
    & 40.77 $\pm$ 0.78
    & 40.29 $\pm$ 0.41
    \\
    S-Trans
    & 44.01 $\pm$ 0.72
    & 76.97 $\pm$ 0.26
    & 45.97 $\pm$ 15.44
    & 79.62 $\pm$ 0.49
    \\
    \midrule
    OCP 
    & 43.27 $\pm$ 0.44
    & 76.76 $\pm$ 0.26
    & 55.78 $\pm$ 0.49
    & 79.49 $\pm$ 0.29
    \\
    STraTS 
    & 44.65 $\pm$ 0.45
    & 76.69 $\pm$ 0.09
    & 55.78 $\pm$ 0.49
    & 79.49 $\pm$ 0.29
    \\
    DueTT 
    & 31.43 $\pm$ 1.32
    & 71.79 $\pm$ 1.24
    & 14.89 $\pm$ 0.50
    & 54.63 $\pm$ 9.00
    \\
    \midrule
    EBCL
    & 46.02 $\pm$ 0.10
    & \textbf{77.23 $\pm$ 0.06}
    & \textbf{55.98 $\pm$ 0.72}
    & \textbf{80.96 $\pm$ 0.11}
    \\
    \bottomrule
\end{tabular}
\end{center}

\caption{\textbf{Linear Probing for Evaluating the Pretrained Embeddings.} EBCL (boldfaced) yielded the embedding that performs the best across all tasks.}

\label{tab:mimic_linear_probe_apr}
\small
\begin{center}
  \begin{tabular}{lccccccc}
  \toprule
    &\multicolumn{2}{c}{MIMIC-IV Hypotension} &\multicolumn{2}{c}{MIMIC-IV Mechanical Ventilation}\\
    \cmidrule(r){2-3}\cmidrule(r){4-5}
    & In-ICU Mortality & 3-Day LOS & In-ICU Mortality & 3-Day LOS \\
    \midrule
    \midrule
    OCP
    & 31.61 ± 1.46 & 68.62 ± 0.93 & 21.60 ± 2.12 & 65.12 ± 0.68
    \\
    STraTS
    & 32.73 ± 2.03 & 69.19 ± 1.20 & 27.96 ± 1.80 & 67.02 ± 0.90
    \\
    DueTT
    & 31.14 ± 1.41 & 72.16 ± 0.58 & 15.13 ± 0.04 & 63.73 ± 0.27
    \\
    \midrule
    EBCL
    & \textbf{38.35 ± 0.94} & \textbf{73.94 ± 0.44} & \textbf{35.00 ± 2.07} & \textbf{73.89 ± 1.03}
    \\
    \bottomrule
\end{tabular}
\end{center}
\end{table*}

\section{Ablations}

\begin{table*}[ht!]
\caption{\textbf{Ablation Studies}. We compare the EBCL variants by applying various definitions of events and different sampling strategies. We summarize the downstream finetuning performance (AUPRC) of three prediction tasks averaged over 5 runs with different seeds, where we boldface statistically significant results. *Note that FT stands for finetuning data and the entry both means both pre and post data are used, but for predicting 1-Week LOS we always only used the Pre-event dataset to avoid data leakage.}
\label{tab:ablation_apr}
\begin{center}
\small
\centering
  \begin{tabular}{lllcccccc}
  \toprule
    & Index Event & FT* & 30-Day Readmission & 1-Year Mortality & 1-Week LOS \\
    \midrule
    EBCL
    & Inpatient
    & Both
    & \textbf{86.40 $\pm$ 0.03}
    & \textbf{90.55 $\pm$ 0.03}
    & \textbf{90.96 $\pm$ 0.04}
    \\
    Censoring
    & Inpatient 
    & Both
    & 85.83	$\pm$ 0.04
    & 90.08	$\pm$ 0.02
    & 90.35	$\pm$ 0.12
    \\
    Non-Adm 
    & Non-Inpatient
    & Both
    & 85.75 $\pm$ 0.05
    & 90.45 $\pm$ 0.06
    & 90.06	$\pm$ 0.18
    \\
    Outpatient
    & Outpatient
    & Both
    & 85.66 $\pm$ 0.11
    & 90.34	$\pm$ 0.02
    & 89.64	$\pm$ 0.12
    \\
    \midrule
    Pre Event
    & Inpatient
    & Pre
    & 85.70 $\pm$ 0.06
    & 88.89 $\pm$ 0.05
    & \ding{55}
    \\
    Post Event
    & Inpatient
    & Post
    & 85.11 $\pm$ 0.06
    & 89.41 $\pm$ 0.01
    & \ding{55}
    \\
    \midrule
\end{tabular}
\end{center}
\end{table*}

\subsection{Effect of Sampling Observations} \label{apd:censor}
\begin{figure*}[t]
\centering
  \includegraphics[width=0.8\textwidth]{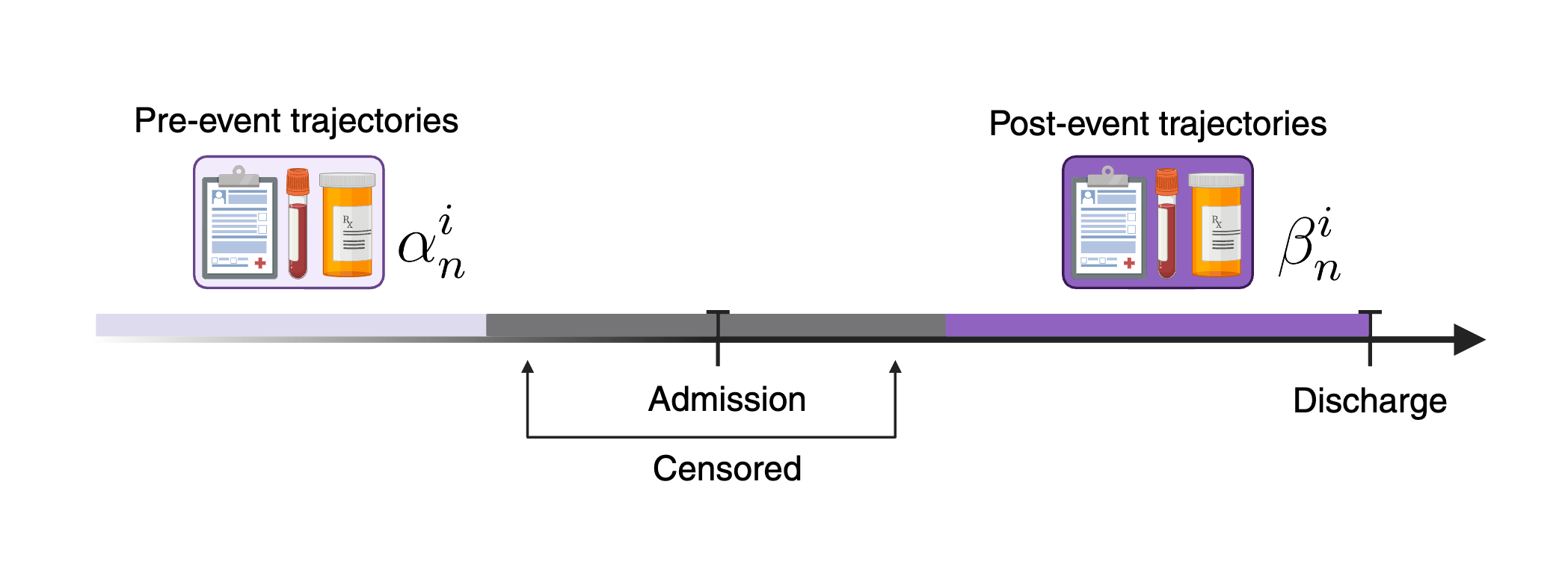}
  \caption{\textbf{EBCL - Censoring Ablation}. We censor a fixed number of observations from the end of the pre window and beginning of the post window and take the 512 remaining closest observations before the admission as pre data and after the admission and post data.} \label{fig:censoring}
\end{figure*}
The censoring window length was determined based on the EBCL window statistics analysis. Specifically, we selected the window to align with the 25th percentile for pre-event and post-event observations, according to the EBCL statistics. Excluding the 25th percentile of data preceding and following the event onset led to the exclusion of 260 observations pre-event and 60 observations post-event. 

\begin{table*}[h]
\caption{\textbf{Results for finetuning only using pre-admission (Pre-event) data}.}
\footnotesize
\label{tab:pre_only}
\begin{center}
  \begin{tabular}{lcccc}
  \toprule
    & \multicolumn{2}{c}{30-Day Readmission} &\multicolumn{2}{c}{1-Year Mortality} \\ 
    \cmidrule(r){2-3}\cmidrule(r){4-5} 
    & AUC & APR  & AUC & APR \\ 
    \midrule
    XGBoost
    & 69.86 $\pm$ 0.03
    & 85.35 $\pm$ 0.02
    & 77.27 $\pm$ 0.04
    & 87.76 $\pm$ 0.02
    \\
    Supervised
    & 68.27 $\pm$ 0.16
    & 84.37 $\pm$ 0.15
    & 77.36 $\pm$ 0.19
    & 87.42 $\pm$ 0.20
    \\
    \midrule
    OCP 
    & 68.83 $\pm$ 0.13
    & 84.63 $\pm$ 0.08
    & 77.93 $\pm$ 0.13
    & 87.85 $\pm$ 0.04
    \\
    STraTS 
    & 68.46 $\pm$ 0.04
    & 84.34 $\pm$ 0.18
    & 78.18 $\pm$ 0.63
    & 87.89 $\pm$ 0.52
    \\
    DueTT 
    & 68.23	$\pm$ 0.11
    & 84.97 $\pm$	0.03
    & 75.14	$\pm$ 0.37
    & 86.17	$\pm$ 0.23
    \\
    \midrule
    EBCL
    & 70.48 $\pm$ 0.09
    & 85.70 $\pm$ 0.06
    & 79.63 $\pm$ 0.06
    & 88.89 $\pm$ 0.05
    \\
    \bottomrule
\end{tabular}
\end{center}
\end{table*}
\begin{table*}[h]
\caption{\textbf{Results for finetuning only using post-admission (Post-event) data.} }
\footnotesize
\label{tab:post}
\begin{center}
  \begin{tabular}{lcccc}
  \toprule
    & \multicolumn{2}{c}{30-Day Readmission} &\multicolumn{2}{c}{1-Year Mortality}\\
    \cmidrule(r){2-3}\cmidrule(r){4-5} 
    & AUC & APR  & AUC & APR \\
    \midrule
    XGBoost
    & 68.05 $\pm$ 0.10
    & 84.80 $\pm$ 0.04
    & 77.12 $\pm$ 0.17
    & 87.54 $\pm$ 0.16
    \\
    Supervised
    & 68.68 $\pm$ 0.08
    & 84.69 $\pm$ 0.05
    & 79.77 $\pm$ 0.21
    & 88.92 $\pm$ 0.19
    \\
    \midrule
    OCP 
    & 68.49 $\pm$ 0.10
    & 84.59 $\pm$ 0.05
    & 79.23 $\pm$ 0.13
    & 88.58 $\pm$ 0.07
    \\
    STraTS 
    & 67.61 $\pm$ 0.16
    & 84.05 $\pm$ 0.10
    & 79.60 $\pm$ 0.12
    & 88.71 $\pm$ 0.10
    \\
    DueTT 
    & 67.47 $\pm$ 0.09
    & 84.14	$\pm$ 0.06
    & 76.47 $\pm$ 0.74
    & 87.15	$\pm$ 0.35
    \\
    \midrule
    EBCL
    & 69.20 $\pm$ 0.09
    & 85.11 $\pm$ 0.06
    & 80.36 $\pm$ 0.03
    & 89.41 $\pm$ 0.01
    \\
    \bottomrule
\end{tabular}
\end{center}
\end{table*}

\section{Risk Stratification and Patient Subtyping with EBCL}
\begin{table*}[h!]
\caption{\textbf{Comparative Analysis of One-Year Mortality and 30-Day Readmission Rates Across Patient Clusters Derived from Different Embedding Strategies.}
Note that the one-year mortality prevalence in the test data is 31.09\%. The 30-Day readmission prevalence is 26.66\%.}
\footnotesize
\label{tab:pheno}
\begin{center}
\centering
  \begin{tabular}{l|ccc|ccc}
  \toprule
    & \multicolumn{3}{c|}{1-Year Mortality (\%)} & \multicolumn{3}{c}{30 Days Readmission (\%) } \\
    \cmidrule(r){2-4}\cmidrule(r){5-7}
    Cluster & Cluster 1 & Cluster 2  & $\Delta$ prevalence & Cluster 1 & Cluster 2 &  $\Delta$ prevalence\\
    \midrule
    OCP
    & 23.46
    & 33.18
    & 9.72
    & 19.86
    & 28.43
    & 8.57
    \\
    STraTS
    & 30.96 
    & 31.64
    & 0.68
    & 26.25
    & 28.75
    & 2.50
    \\
    DueTT
    & 28.07 
    & 34.16
    & 6.09
    & 25.09
    & 27.96
    & 2.87
    \\
    EBCL 
    & 28.00 
    & 50.87
    & 22.87
    & 23.52
    & 34.53
    & 11.01
    \\
    \bottomrule
\end{tabular}
\end{center}
\end{table*}

\subsection{Deep Clustering of EBCL Embedding} 
\label{apd:cluster}

From the EBCL pretrained transformer, we build pre-event and post-event vector representations for each admission in the test set to form a concatenated transformer embedding that includes comprehensive information pre- and post-event. For clustering our representations, we use $K$-means clustering and find the optimal $K$ using the elbow method. This entails plotting within-cluster similarity vs the number of clusters and using an elbow detection method \citep{kneedle} to find the point of maximum curvature which is the optimal $K=6$ for EBCL embeddings.

We use dimensional reduction with Uniform Manifold Approximation and Projection (UMAP) \citep{mcinnes2020umap} for the ease of visualization of our high-dimensional data. The identified clusters that were found are superimposed over the resulting UMAP graphic to highlight the separation and distribution of the clusters in the reduced-dimensional space in Figure \ref{fig:cluster} (a). To understand the phenotype of each cluster, we compare the outcome of patients in each identified cluster (years to mortality, days to readmission, the prevalence of 1-year mortality, and 30-day readmission) (Figure \ref{fig:cluster}). To further validate the prognostic differentiation between the identified heart failure subgroups, we plotted Kaplan-Meier (KM)~\citep{dudley2016introduction} survival curves of the patients in each cluster in figure \ref{fig:cluster} (b). This approach allowed us to visually and statistically compare the distribution of time-to-event of patients in the clusters.

\subsection{Risk Stratification}
We initialized the model with the pretrained model weights of all baseline models (EBCL, OCP \citep{agrawal2022leveraging}, STraTS \citep{tipirneni2022self}, DuETT \citep{labach2023DuETT}) and concatenated the pre- and post-event embeddings of the test set patients. We applied UMAP \citep{mcinnes2020umap} to reduce the dimensionality of the embeddings and to easily visualize our high dimensional data. Furthermore, we applied K-means clustering with $k=2$ on the pretrained embeddings to subgroup the patient cohort by outcome risk. Using the identified clusters, we evaluated the survival outcomes with the Kaplan-Meier estimator \citep{dudley2016introduction}, which provided a visual representation of the survival probability over time for each cluster. To quantitatively compare the prognosis of two identified clusters (Cluster 1 and 2), we compare days to mortality and days to readmission between Cluster 1 and Cluster 2 (Figure \ref{fig:cluster}, Table \ref{tab:pheno}).

\begin{figure*}[h!]
\centering
  \includegraphics[width=\textwidth]{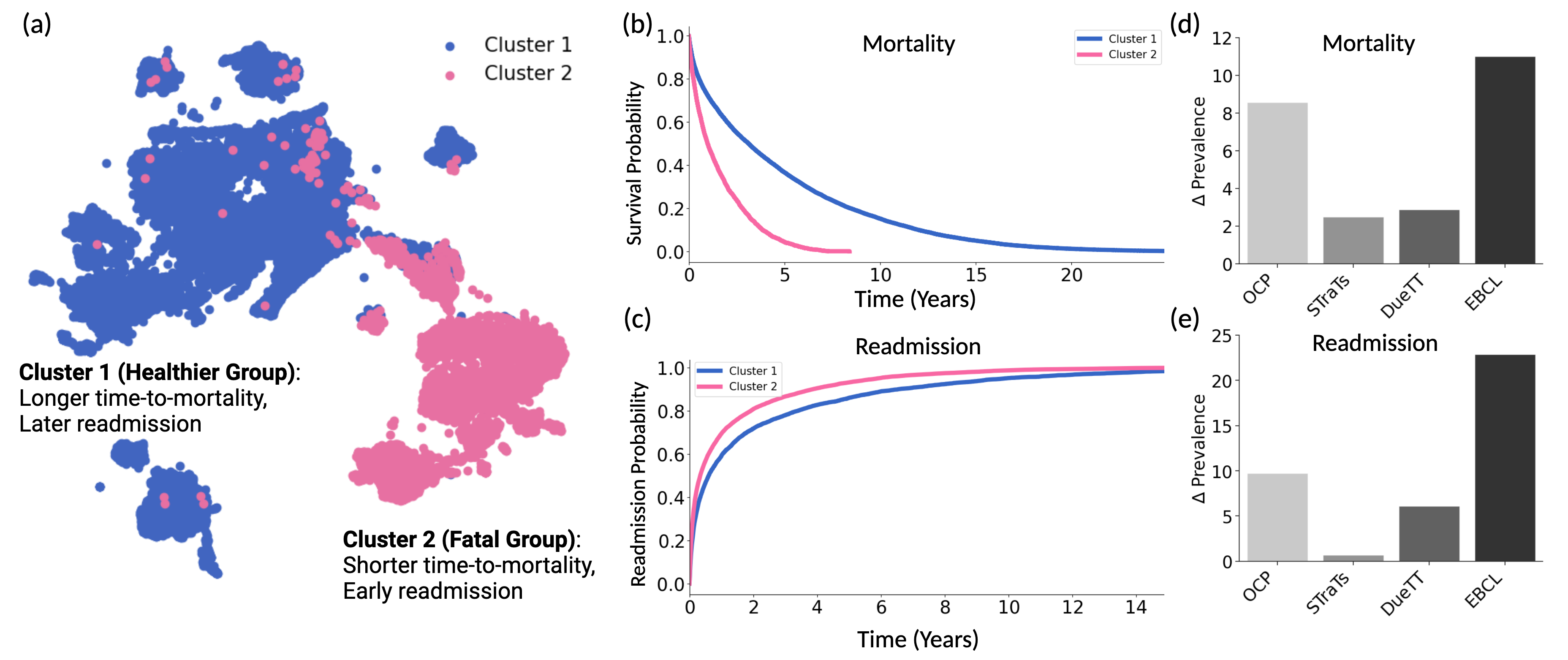}
  \caption{\textbf{Clustering and outcome risk subtyping from pretrained EBCL Embedding.} \textbf{(a)} Clustering and dimensionality reduction of pretrained embedding showed distinct heart failure clusters with unique outcome prognoses. Clusters are sorted by prevalence of 1-Year Mortality in each, thus cluster 1 has the highest 1-Year Mortality and represents the sickest group, and cluster 6 represents the healthiest. \textbf{(b)} Mortality survival curve of the same clusters from Figure (a).  \textbf{(c)} Plot of the proportion of population readmitted over time for the same clusters from Figure (a).}
  \label{fig:cluster}
\end{figure*}

Our analysis revealed statistically significant disparities in outcome prevalence among the clusters identified by various predictive models, including EBCL, OCP \citep{agrawal2022leveraging}, and STraTS \citep{tipirneni2022self}. Specifically, using a two-sampled Student's t-test, we found that Cluster 2 in the EBCL model demonstrated a statistically significant higher one-year mortality rate at 50.87 compared to 28.00 in Cluster 1 (Table \ref{tab:pheno}). This trend was consistent across other models, with the EBCL model showing the highest gaps in prevalence, reinforcing its robustness in stratifying patient risk. However, the clusters identified with STraTS \citep{tipirneni2022self} on the 1-Year Mortality task didn't show patient subgroups that are with statistically significantly different prognoses. 

Notably, the EBCL model showed better stratification of patients, evidenced by high gaps of prevalence between clusters compared to other baselines (22.87 for one-year mortality and 11.01 for 30-day readmission tasks, respectively. (Table \ref{tab:pheno}, $\Delta$ prevalence column), Figure \ref{fig:cluster} (d), (e)). These insights demonstrate the capacity of EBCL to generate clinically meaningful representations, potentially aiding in more effective patient stratification and personalized healthcare based on patient phenotyping.

\end{document}